\documentclass[10pt,journal,compsoc]{IEEEtran}
\pdfoutput=1
\usepackage[pagebackref=true,breaklinks=true,letterpaper=true,colorlinks,citecolor=blue,linkcolor=blue,bookmarks=false]{hyperref}
\usepackage{array}
\usepackage{arydshln}
\usepackage{colortbl}
\usepackage{makecell}
\usepackage{ragged2e}

\usepackage{xspace}
\usepackage{color}
\usepackage{multirow}
\usepackage{graphics} 
\usepackage{adjustbox}
\usepackage{amsmath}
\usepackage{amssymb}  
\usepackage{paralist} 
\usepackage{enumitem} 
\usepackage{booktabs} 

\graphicspath{{images/}}

\long\def\ignorethis#1{}

\definecolor{gray}{rgb}{0.35,0.35,0.35}
\definecolor{MyBlue}{rgb}{0,0.2,0.8}
\definecolor{MyRed}{rgb}{0.8,0.2,0}
\definecolor{MyGreen}{rgb}{0.0,0.5,0.1}
\definecolor{MyGray}{rgb}{0.4,0.4,0.4}

\def\red#1{\textcolor{red}{#1}}
\def\blue#1{\textcolor{blue}{#1}}

\def\first#1{\red{\textbf{#1}}}
\def\second#1{\blue{\underline{#1}}}

\newlength\paramargin
\newlength\figmargin
\newlength\secmargin

\usepackage{array}
\newcolumntype{L}[1]{>{\raggedright\let\newline\\\arraybackslash\hspace{0pt}}m{#1}}
\newcolumntype{C}[1]{>{\centering\let\newline\\\arraybackslash\hspace{0pt}}m{#1}}
\newcolumntype{R}[1]{>{\raggedleft\let\newline\\\arraybackslash\hspace{0pt}}m{#1}}

\newcommand{\secref}[1]{Section~\ref{sec:#1}}
\newcommand{\figref}[1]{Figure~\ref{fig:#1}}
\newcommand{\tabref}[1]{Table~\ref{tab:#1}}

\ifCLASSOPTIONcompsoc
\usepackage[nocompress]{cite}
\else
\usepackage{cite}
\fi

\ifCLASSINFOpdf
\else
\fi
\begin{document}
\title{ITSRN++: Stronger and Better Implicit 
Transformer Network for Continuous Screen Content Image Super-Resolution}

\author{Sheng Shen,
Huanjing Yue,
Jingyu Yang,
Kun Li
\IEEEcompsocitemizethanks{
\IEEEcompsocthanksitem S. Shen, H. Yue, and J. Yang are with the School of Electrical and Information Engineering, Tianjin University, Tianjin China. 
\IEEEcompsocthanksitem K. Li is with the College of Intelligence and Computing, Tianjin University, Tianjin, China. 
%
}
}

\markboth{}%
{Shell \MakeLowercase{\textit{et al.}}: Bare Demo of IEEEtran.cls for Computer Society Journals}
%



\IEEEtitleabstractindextext{%
\begin{abstract}
\justifying
Nowadays, online screen sharing and remote cooperation are becoming ubiquitous. However, the screen content may be downsampled and compressed during transmission, while it may be displayed on large screens or the users would zoom in for detail observation at the receiver side. Therefore, developing a strong and effective screen content image (SCI) super-resolution (SR) method is demanded. We observe that the weight-sharing upsampler (such as deconvolution or pixel shuffle) could be harmful to sharp and thin edges in SCIs, and the fixed scale upsampler makes it inflexible to fit screens with various sizes. To solve this problem, we propose an implicit transformer network for continuous SCI SR (termed as ITSRN++). Specifically, we propose a modulation based transformer as the upsampler, which modulates the
pixel features in discrete space via a periodic nonlinear
function to generate features for continuous pixels. To enhance the extracted features, we further propose an enhanced transformer as the feature extraction backbone, where convolution and attention branches are utilized parallelly. Besides, we construct a large scale SCI2K dataset to facilitate the research on SCI SR. Experimental results on nine datasets demonstrate that the proposed method achieves state-of-the-art performance for SCI SR (\textit{outperforming SwinIR by 0.74 dB for $\times3$ SR}) and also works well for natural image SR. Our codes and dataset will be released upon the acceptance of this work.

%
\end{abstract}

}

\maketitle

\IEEEdisplaynontitleabstractindextext

%
\IEEEpeerreviewmaketitle


\IEEEraisesectionheading{\section{Introduction \label{sec:introduction} }}

%
%

\IEEEPARstart{S}{creen} content images (SCIs)~\cite{peng2016screen_overview}, which refer to the contents generated or rendered by computers, such as graphics and texts, are becoming popular due to the widely used screen sharing, remote cooperation, and online education. However, the images may be downsampled and compressed during transmission due to limited bandwidth. Meanwhile, the received images may be displayed on a large screen and users may zoom in the image for detail observation. Therefore, image super-resolution (SR) is demanded to improve the quality of SCIs.

\begin{figure*}
	\newlength\fs
	\setlength{\fs}{-0.4cm}
	\scriptsize
	\centering
    \resizebox{0.98\textwidth}{!}{
	\begin{tabular}{lccccccc}
	
	 Scale & \hspace{\fs} Resolution & Bicubic &  MetaSR~\cite{hu2019metasr} & LIIF\cite{chen2021liif} & ITSRN~\cite{yang2021itsrn}  &  LTE~\cite{lee2021lte} & ITSRN++ 
	 \\
	 \hline
	 
	 \\
	 \hspace{-0.24cm}
	    
	 \begin{tabular}{c}
    $\times 1.0 $
        \\
	 \end{tabular}
	  & \hspace{\fs} {$ 180 \times 320$}
	 & \hspace{\fs}{\includegraphics[width=0.10\textwidth]{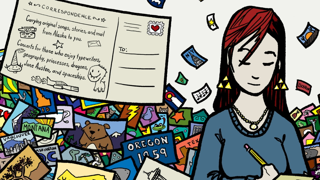}}
	 & \hspace{\fs}{\includegraphics[width=0.10\textwidth]{Image/result/x1.0/SCI10_lr.png}}
	 & \hspace{\fs}{\includegraphics[width=0.10\textwidth]{Image/result/x1.0/SCI10_lr.png}}
	 
	 & \hspace{\fs}{\includegraphics[width=0.10\textwidth]{Image/result/x1.0/SCI10_lr.png}}
	 & \hspace{\fs}{\includegraphics[width=0.10\textwidth]{Image/result/x1.0/SCI10_lr.png}}
	 & \hspace{\fs}{\includegraphics[width=0.10\textwidth]{Image/result/x1.0/SCI10_lr.png}}
	 
	 \\
	 $\times 2.1 $ & \hspace{\fs}  $ 378 \times 672 $
	 & \hspace{\fs}\includegraphics[width=0.10\textwidth]{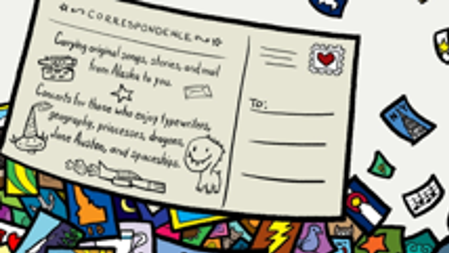}
	 & \hspace{\fs}\includegraphics[width=0.10\textwidth]{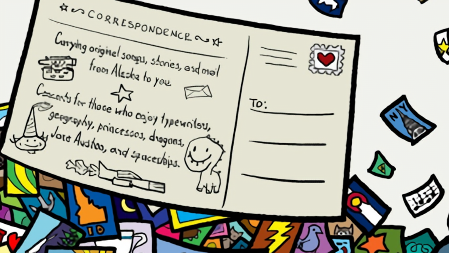}
	 & \hspace{\fs}\includegraphics[width=0.10\textwidth]{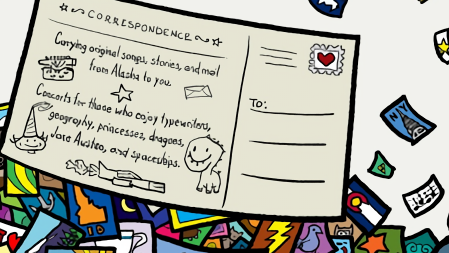}
	 & \hspace{\fs}\includegraphics[width=0.10\textwidth]{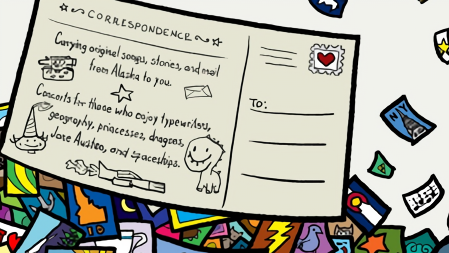}
	 &\hspace{\fs} \includegraphics[width=0.10\textwidth]{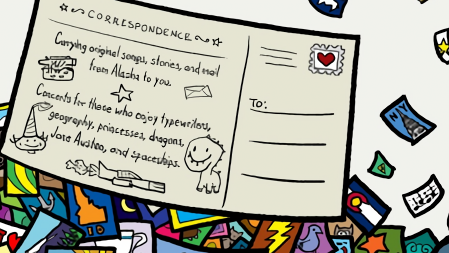}
	 &\hspace{\fs}\includegraphics[width=0.10\textwidth]{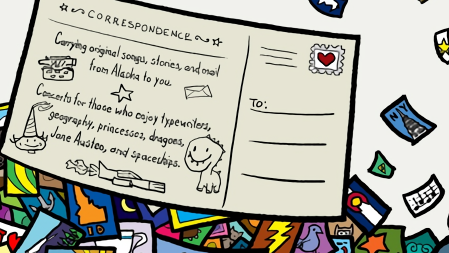}
	 
	 \\
	 $\times 4.2 $ & \hspace{\fs}  $ 756 \times 1344 $
	 &\hspace{\fs} \includegraphics[width=0.10\textwidth]{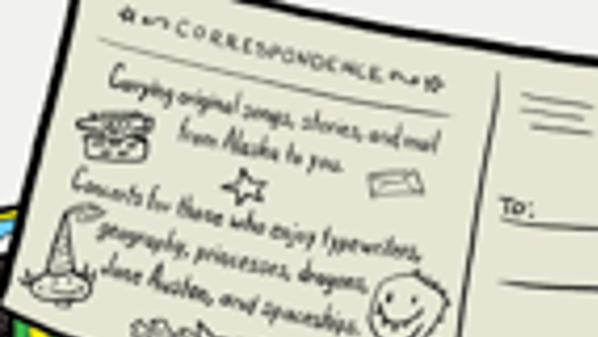}
	 &\hspace{\fs} \includegraphics[width=0.10\textwidth]{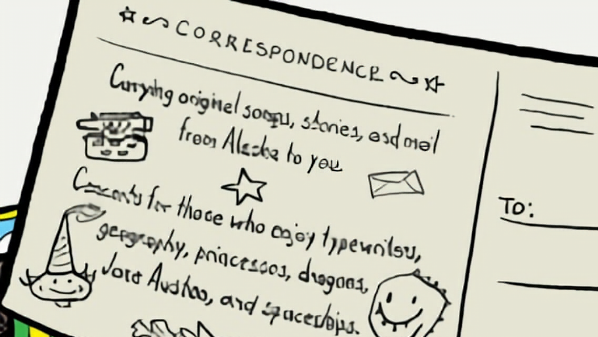}
	 &\hspace{\fs} \includegraphics[width=0.10\textwidth]{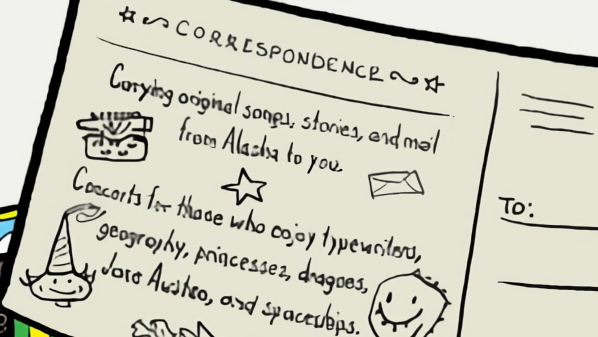}
	 &\hspace{\fs} \includegraphics[width=0.10\textwidth]{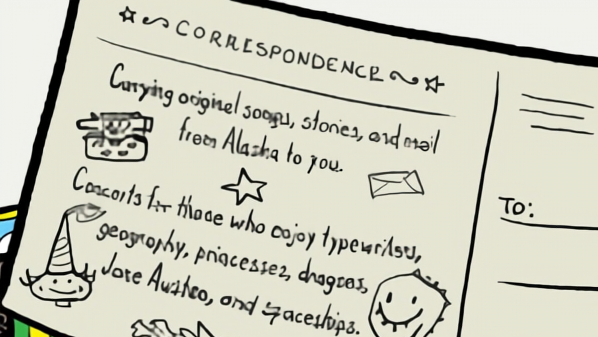}
	 &\hspace{\fs} \includegraphics[width=0.10\textwidth]{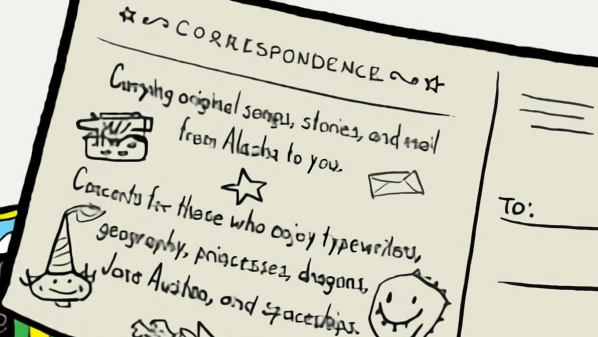}
	 &\hspace{\fs} \includegraphics[width=0.10\textwidth]{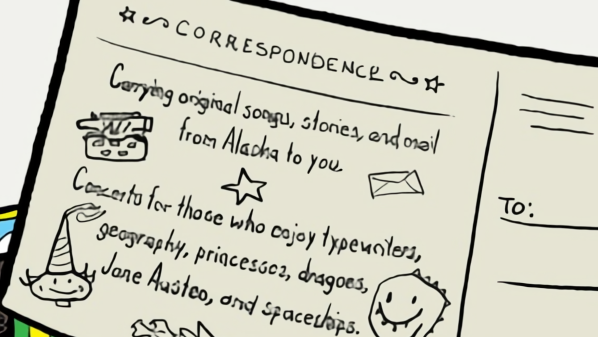}
	 
	 \\	
	 $\times 8.7 $ &  \hspace{\fs} $ 1565 \times 2784 $
	 &\hspace{\fs} \includegraphics[width=0.10\textwidth]{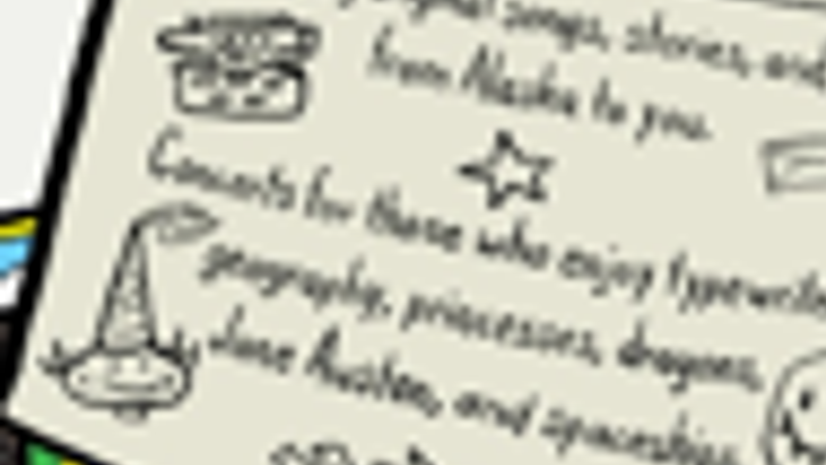}
	 &\hspace{\fs} \includegraphics[width=0.10\textwidth]{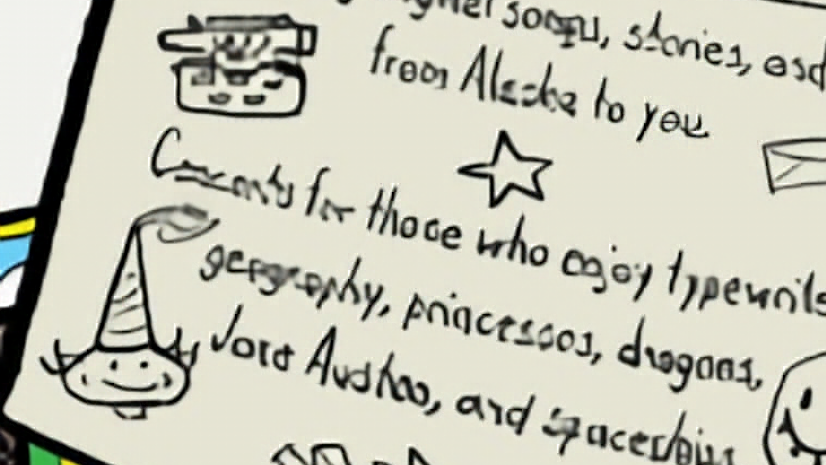}
	 &\hspace{\fs} \includegraphics[width=0.10\textwidth]{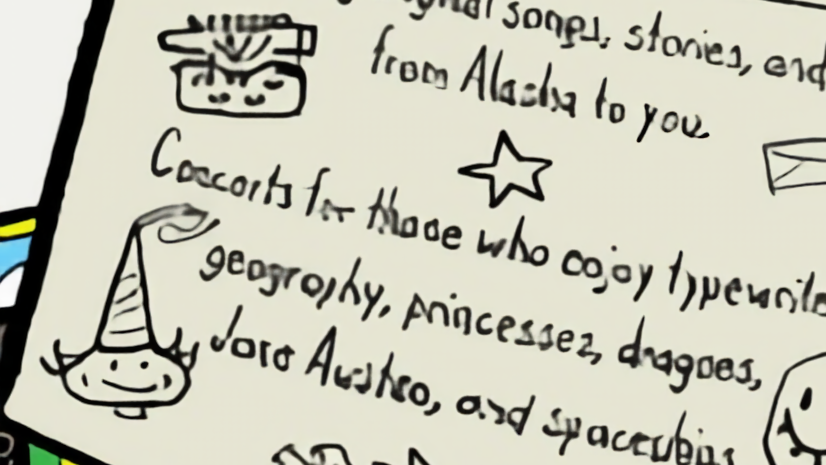}
	 &\hspace{\fs} \includegraphics[width=0.10\textwidth]{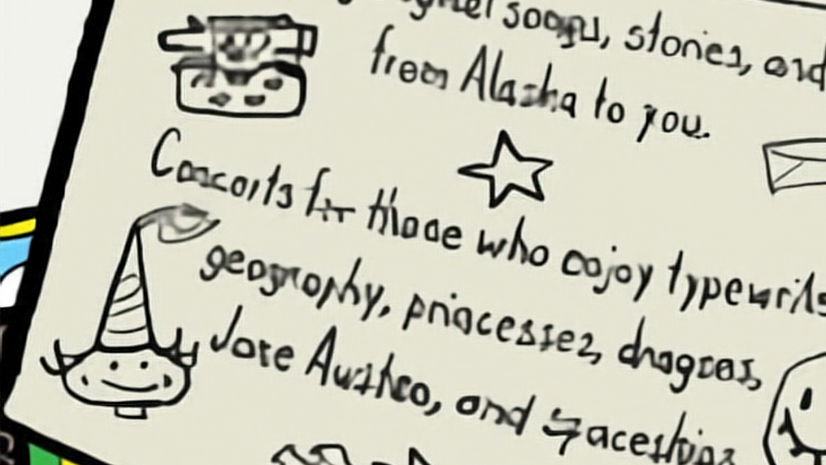}
	 &\hspace{\fs} \includegraphics[width=0.10\textwidth]{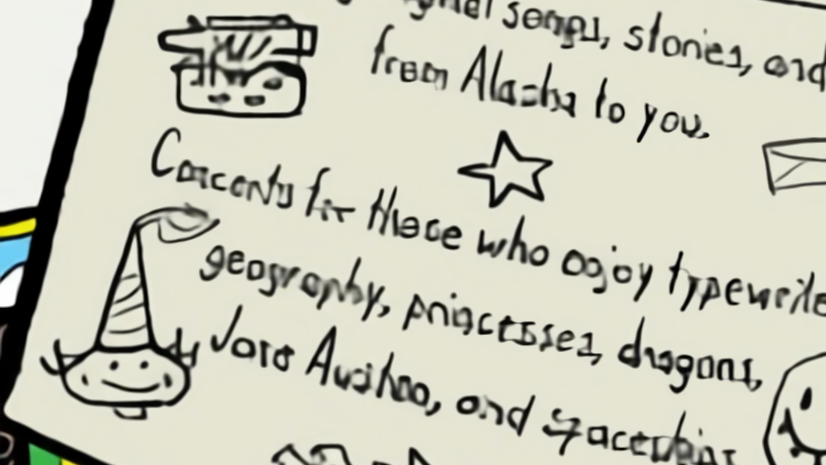}
	 &\hspace{\fs} \includegraphics[width=0.10\textwidth]{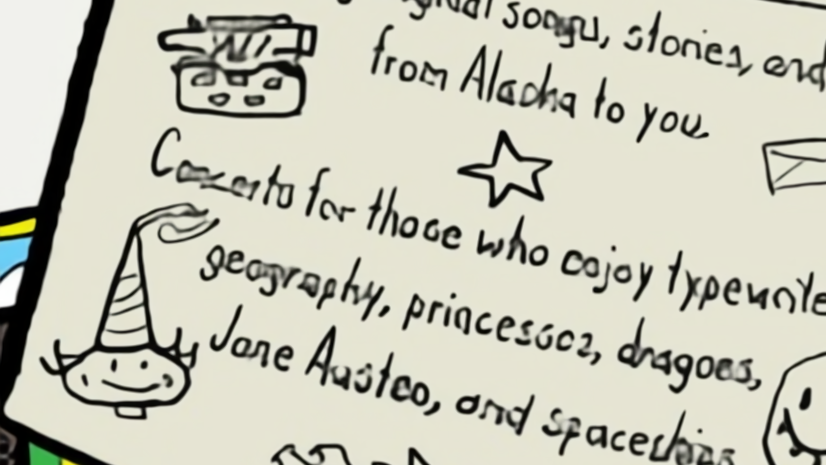}
	 
	 \\	
	 $\times 16.3 $ &  \hspace{\fs} $ 2934 \times 5216 $
	 &\hspace{\fs} \includegraphics[width=0.10\textwidth]{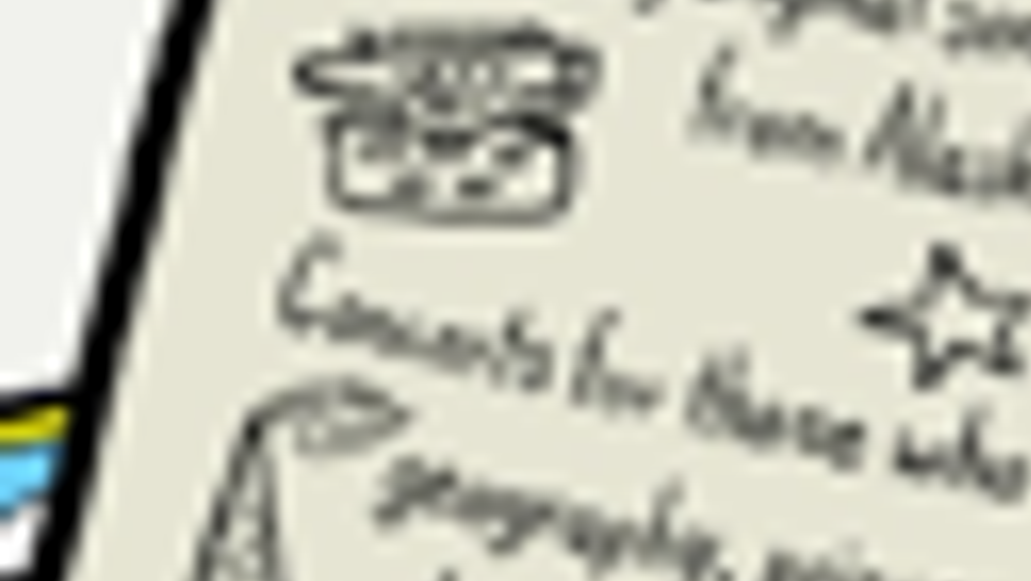}
	 &\hspace{\fs} \includegraphics[width=0.10\textwidth]{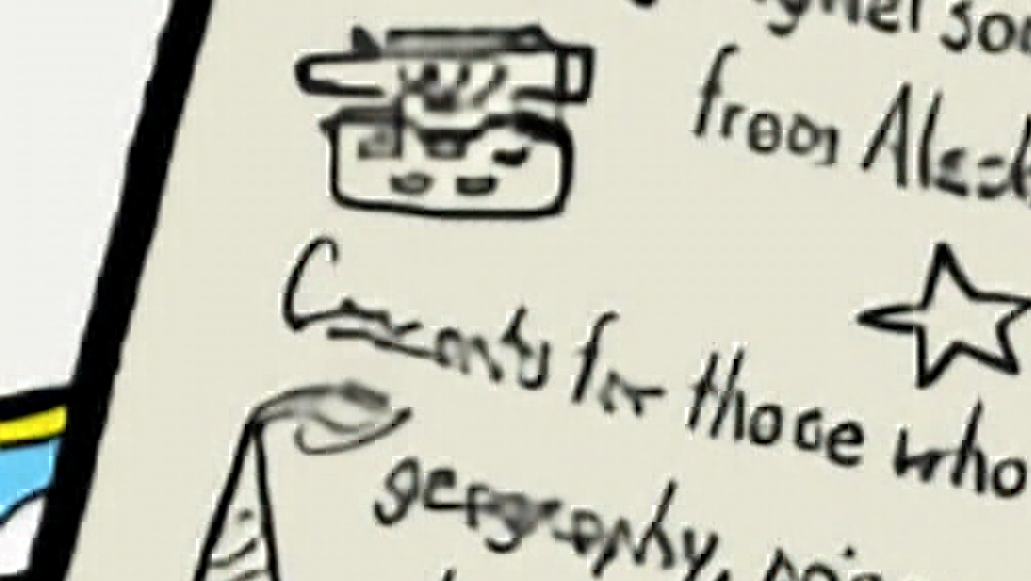}
	 &\hspace{\fs} \includegraphics[width=0.10\textwidth]{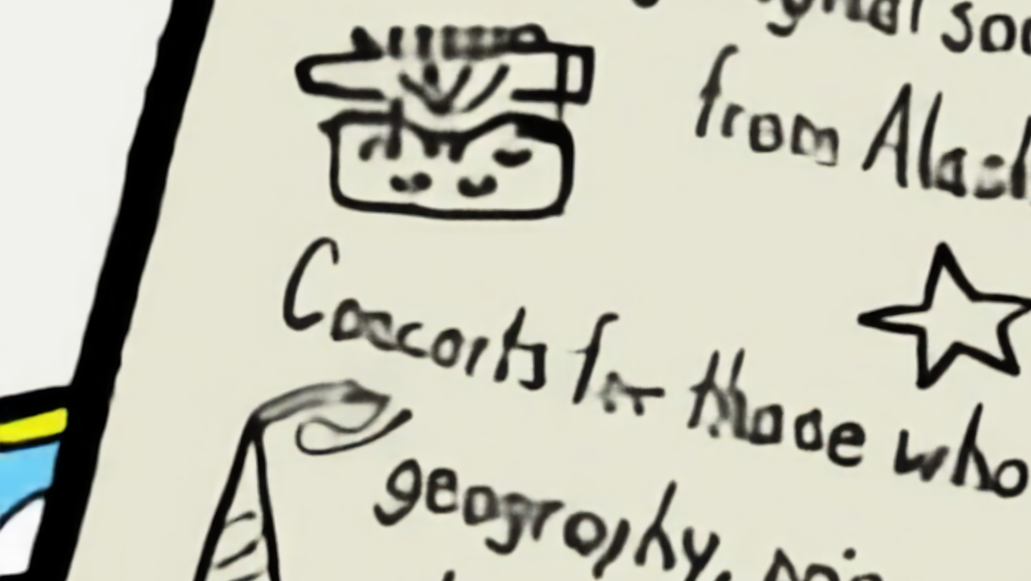}
	 & \hspace{\fs} \includegraphics[width=0.10\textwidth]{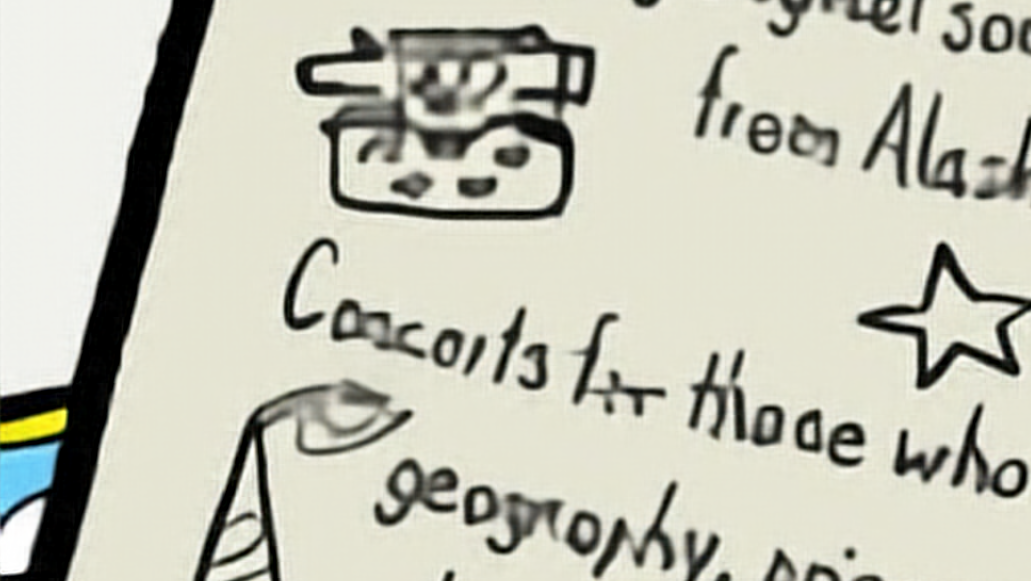}
	 &\hspace{\fs} \includegraphics[width=0.10\textwidth]{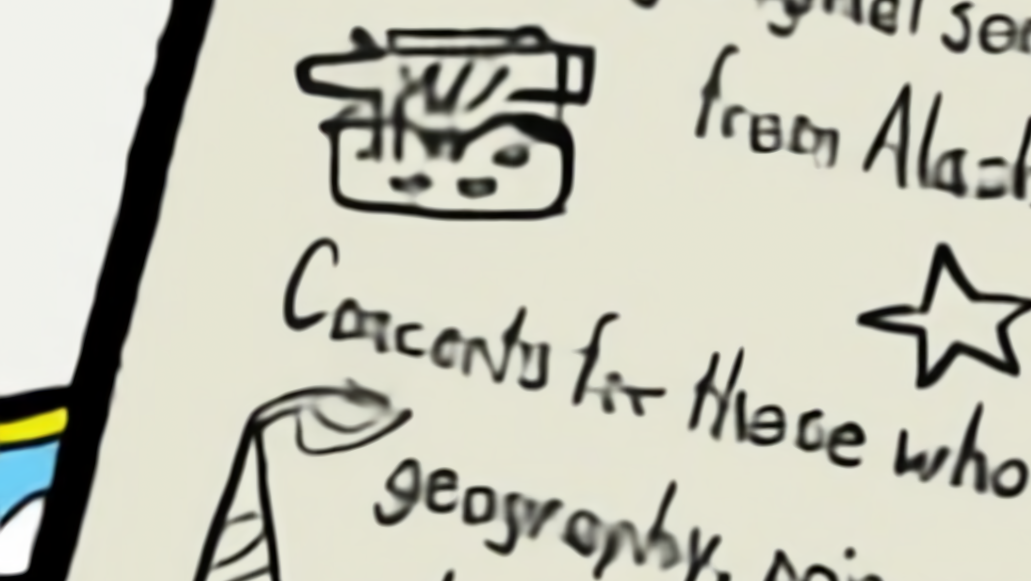}
	 &\hspace{\fs} \includegraphics[width=0.10\textwidth]{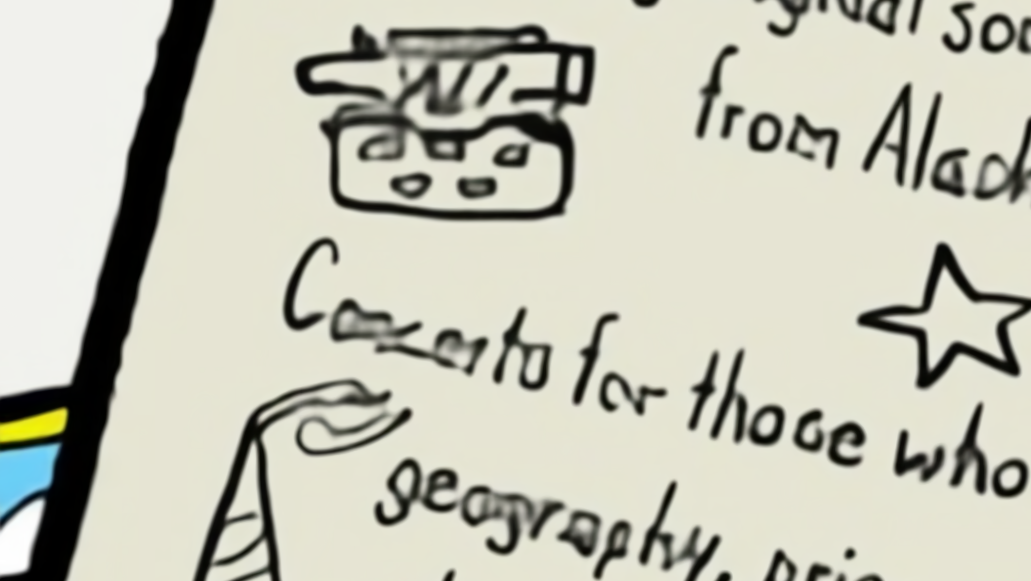}
	 
	 \\
	 $\times 32.5 $ &  \hspace{\fs} $ 5850 \times 10400 $
	 &\hspace{\fs} \includegraphics[width=0.10\textwidth]{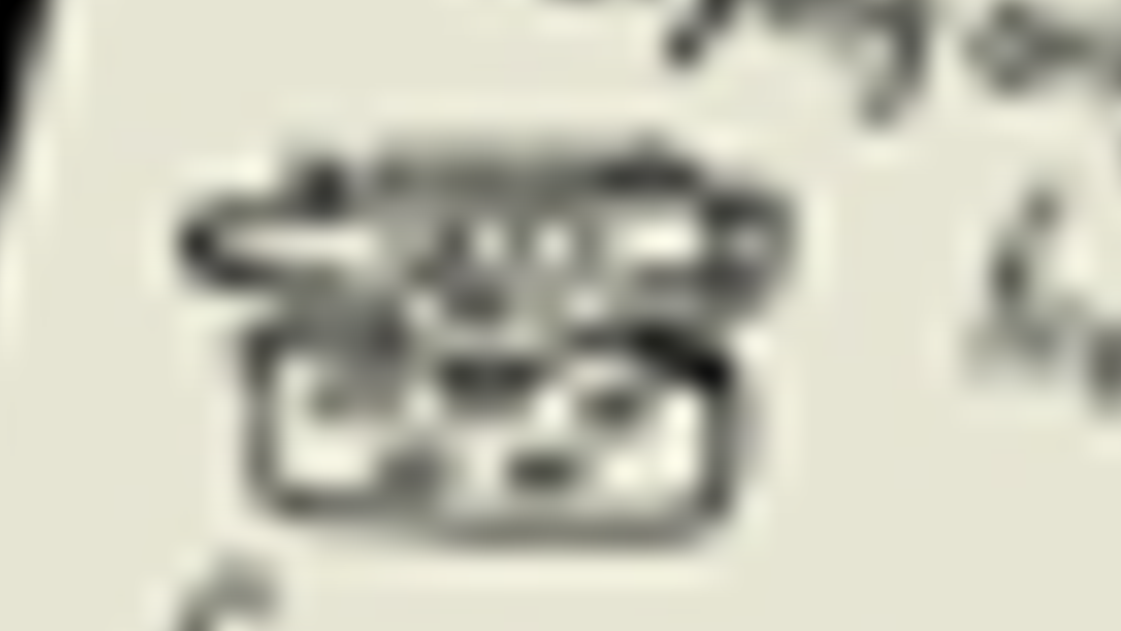}
	 &\hspace{\fs} \includegraphics[width=0.10\textwidth]{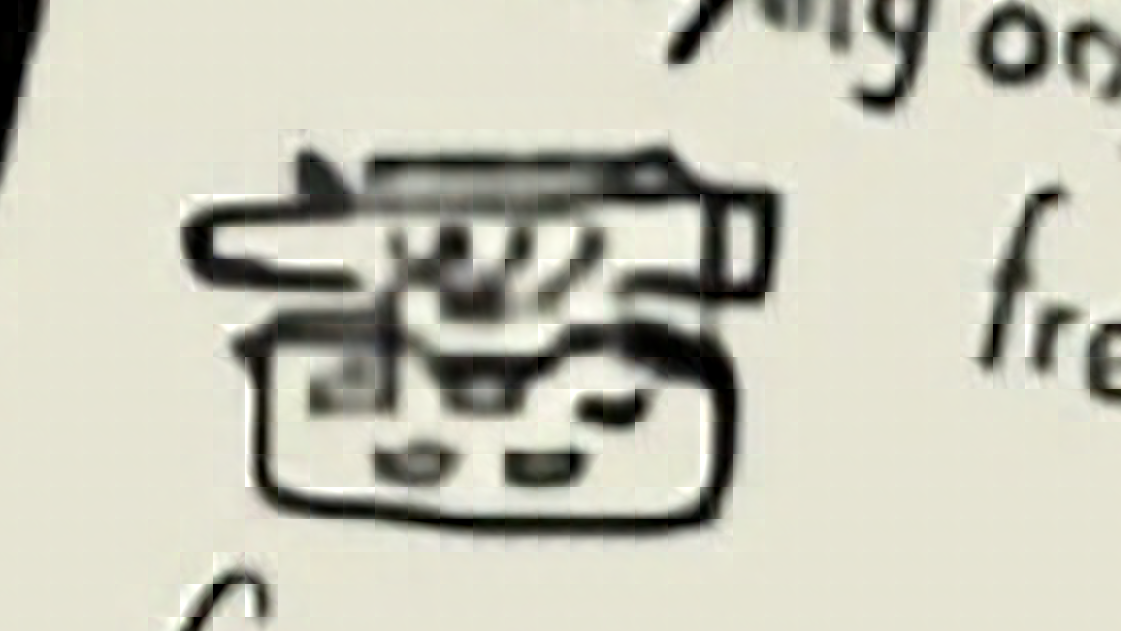}
	 &\hspace{\fs} \includegraphics[width=0.10\textwidth]{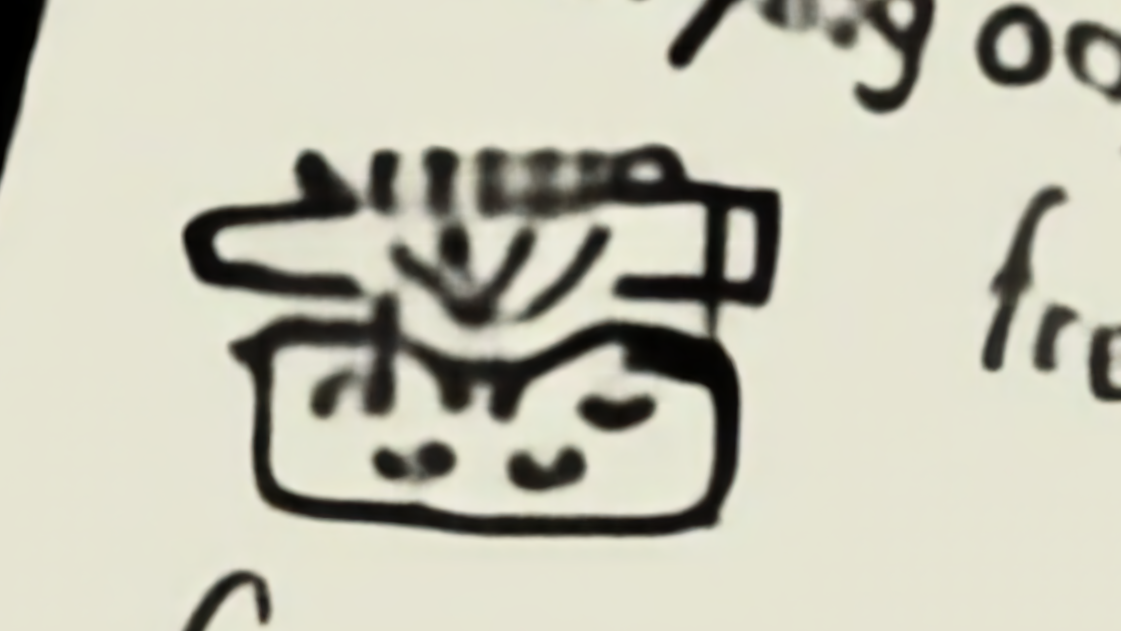}
	 &\hspace{\fs} \includegraphics[width=0.10\textwidth]{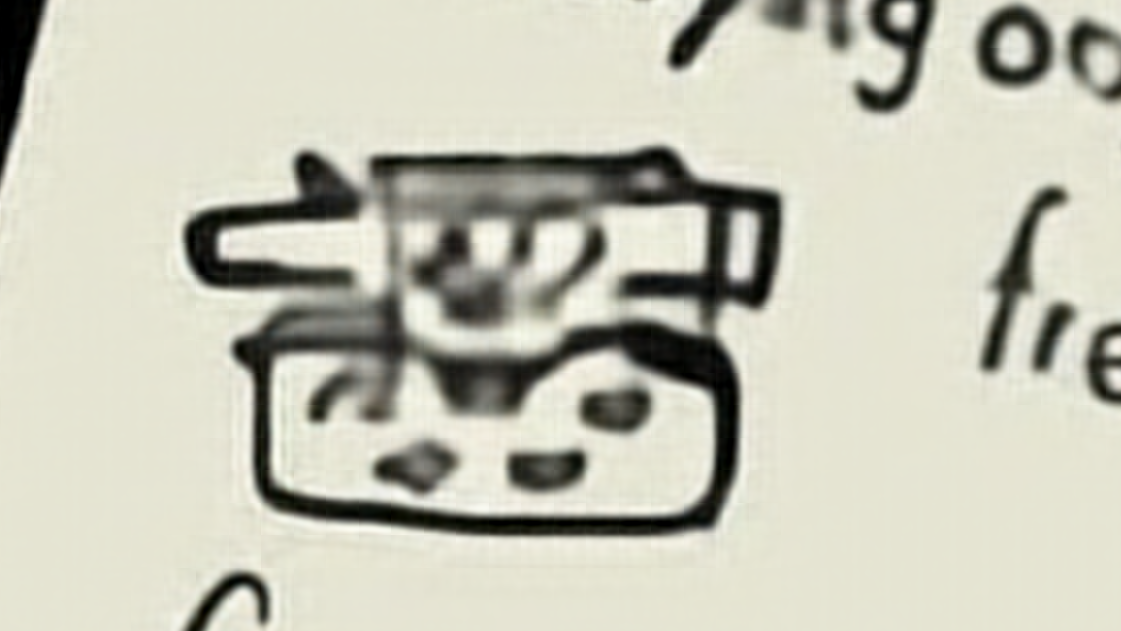}
	 &\hspace{\fs} \includegraphics[width=0.10\textwidth]{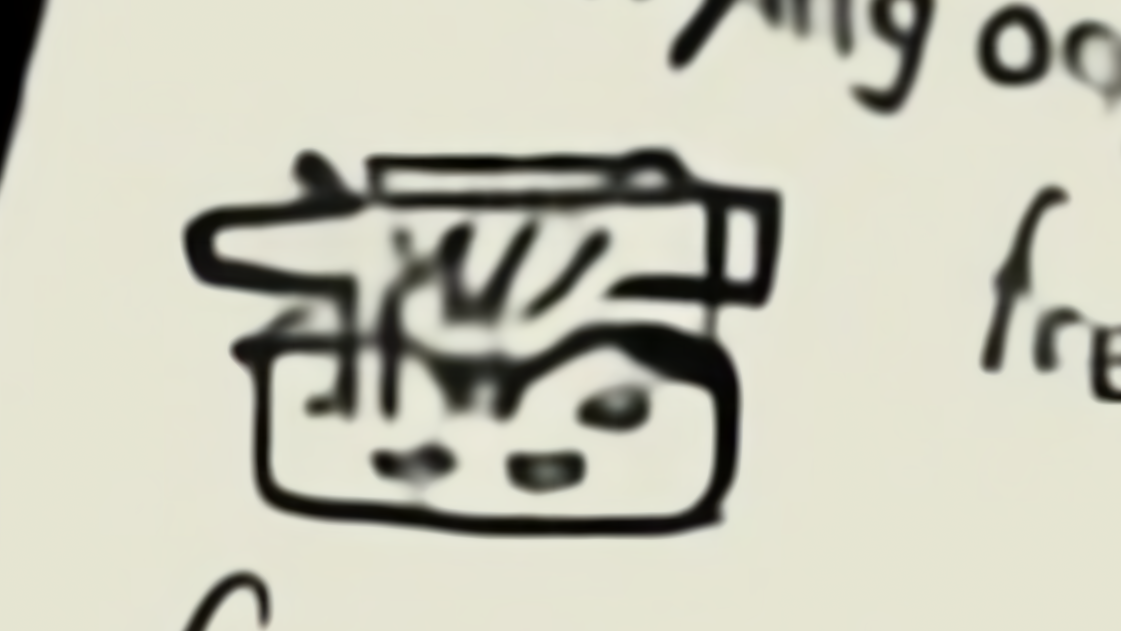}
	 &\hspace{\fs} \includegraphics[width=0.10\textwidth]{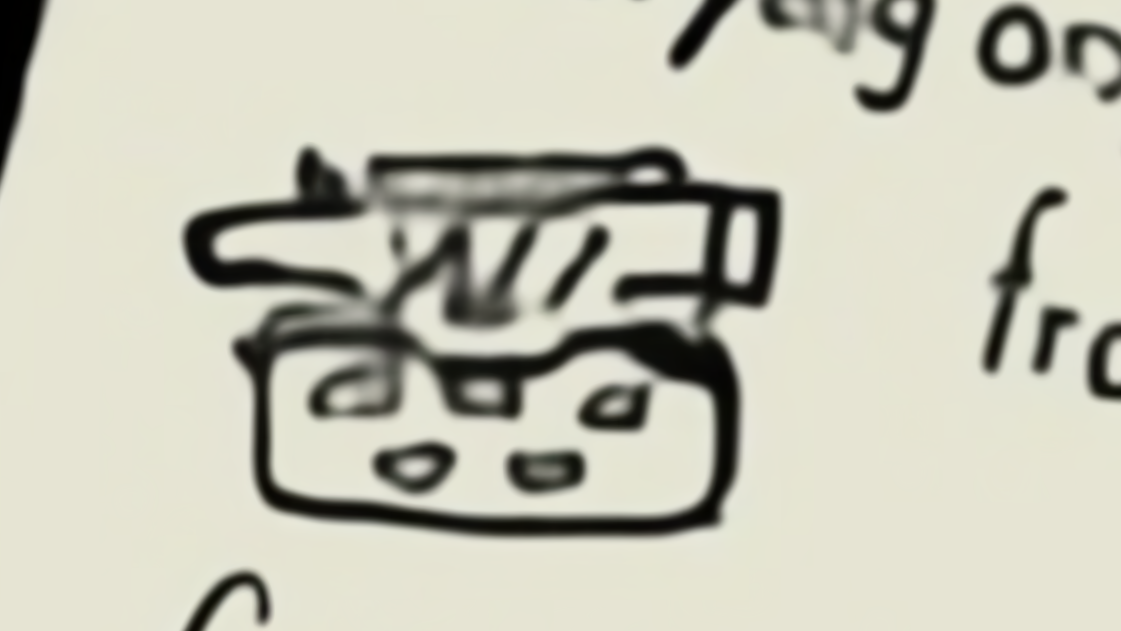}
	
	 \\	 
	 $\times 64.7 $ & \hspace{\fs} $ 11646 \times 20704 $
	 &\hspace{\fs} \includegraphics[width=0.10\textwidth]{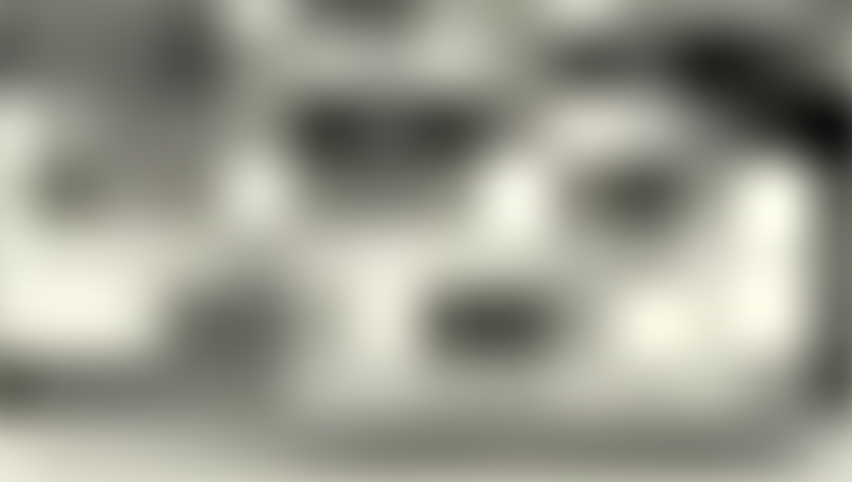}
	 &\hspace{\fs} \includegraphics[width=0.10\textwidth]{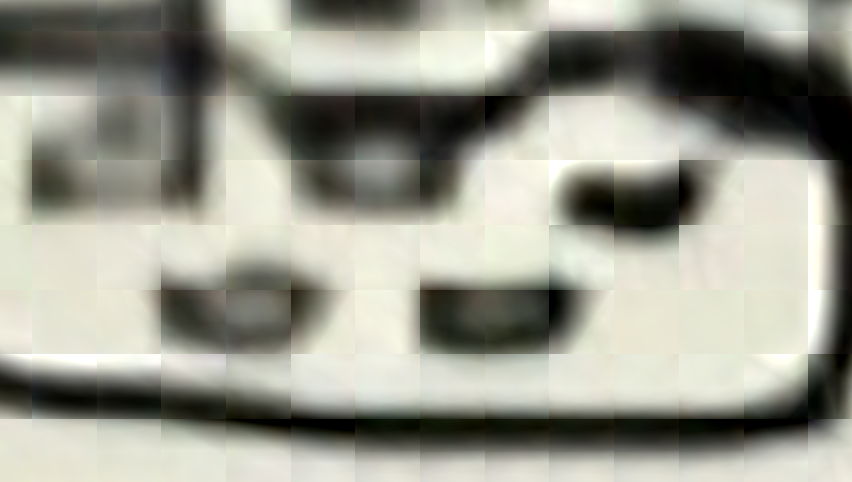}
	 &\hspace{\fs} \includegraphics[width=0.10\textwidth]{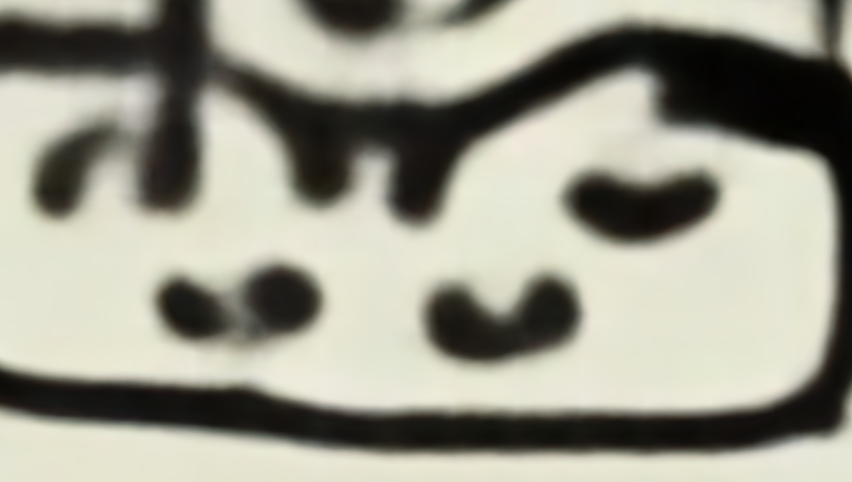}
	 &\hspace{\fs} \includegraphics[width=0.10\textwidth]{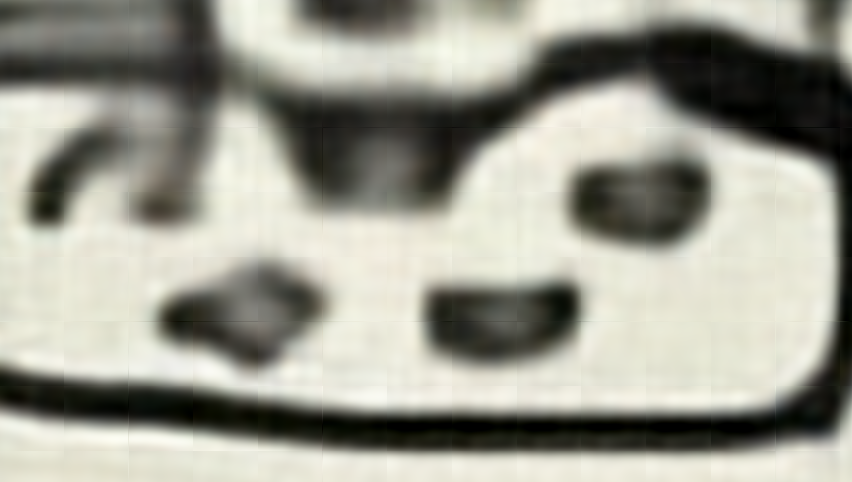}
	 &\hspace{\fs} \includegraphics[width=0.10\textwidth]{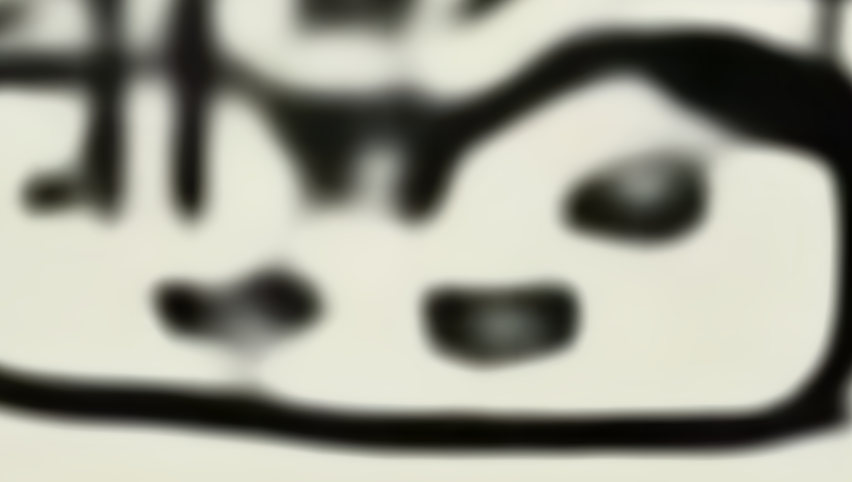}
	 &\hspace{\fs} \includegraphics[width=0.10\textwidth]{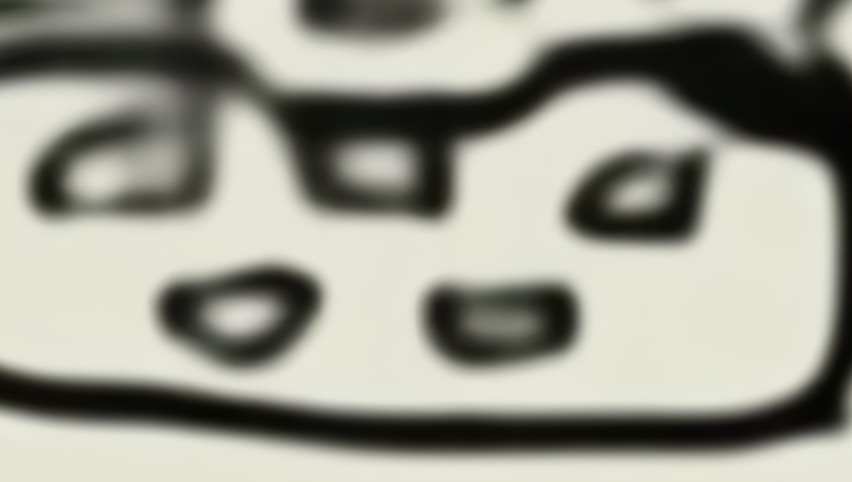}

	 \\	 
	 
	\end{tabular}
    }
	\caption{
		\textbf{Visual comparison of the proposed ITSRN++ with state-of-the art continuous magnification methods. With continuous upsamplers, images can be magnified with arbitrary ratios.}
	}
	\label{fig:arbi_compare}
\end{figure*}

As shown in \figref{arbi_compare}, the SCIs are dominated by sharp edges and high contrast, which makes them different from natural images. We observe that the upsampling module (such as deconvolution or sub-pixel convolution) in most SR networks could be harmful to sharp and thin edges in SCIs since the weight-sharing strategy tend to produce smooth reconstruction results. In addition, the fixed upsampling ratios make them inflexible to fit screens of various sizes. Therefore, developing a continuous upsampler while being friendly to sharp edges is demanded. On the other hand, natural image SR methods \cite{lim2017edsr,zhang2018RDN, zhang2018rcan,zhang2019RNAN,chen2021IPT,li2019SRFBN,liang21swinir,kong2022reflash,lin2022revisiting} are widely explored, while SCI SR is rarely studied. Wang \textit{et al.} \cite{wang2021super} proposed an SR method for compressed screen content videos, which addressed the compression artifacts of screen content videos. However, there are no tailored modules specifically designed for screen content. Developing effective feature extraction backbones for sharp edges also needs to be explored.   
In this work, we propose an implicit transformer based upsampler and enhanced transformer based feature extraction backbone to solve the two problems. In the following, we give motivations for the two modules.

%

\textbf{Upsampler.} 
Previous SR methods usually utilize deconvolution \cite{zeiler2010deconvolutional} or pixel-shuffle (also known as sub-pixel) layer \cite{shi2016subpixel} to serve as the upsampler. However, they are restricted to fixed and integer upsampling ratios. In order to achieve arbitrary SR, Hu \textit{et. al.}\cite{hu2019metasr} proposed meta upsampler to utilize the relative position offset between the HR coordinate and the original LR coordinate to predict the meta weight and then the extracted features are multiplied by the meta weight to generate the SR pixels. 
Different from \cite{hu2019metasr}, Chen \textit{et.al} \cite{chen2021liif} concatenated the relative offsets and the LR features, which go through the multi-layer perceptron (MLP) to generate the SR pixel. We observe that all the upsamplers can be summarized as three steps. 
1) \textbf{Coordinate Projection}. Projecting target coordinates to the LR local space to get relative offsets.  The deconvolution and sub-pixel convolution learn relative offsets implicitly while continuous upsamplers calculate the relative offsets explicitly.
2) \textbf{Weight Generation}. Different upsamplers adopt different weight generation methods and it is the essential part for SR quality. The weights in deconvolution and sub-pixel convolution based upsamplers are fixed while scalable in continuous upsamplers. 
3) \textbf{Aggregation}. Aggregating the features extracted from the LR input according to the predicted weights to generate the SR values. 

Coincidentally, the three steps can be further linked with the transformer $I_q=\Phi(Q,K)V$, where the coordinate projection is the query-key matching process, weight generation is modeled by $\Phi(\cdot,\cdot)$, and aggregation is the multiplication process. Since the input queries are coordinates while the outputs are pixel values, similar to the implicit function defined in Nerf~\cite{2020NeRF}, we term the upsampler transformer as implicit transformer. In this work, we model the features in continuous space (i.e., $V_q$) as the modulated version of features (i.e., $V$) in discrete space, and then $V_q$ is mapped to $I_q$ via a multi-layer perceptron (MLP), as shown in \figref{net_arch}. Specifically, we modulate the pixel feature $V$ via a periodic modulated implicit transformer. In this way, we can generate SR results with arbitrary magnification ratios, as shown in \figref{arbi_compare}.


\textbf{Feature extraction.}
The quality of super-resolved pixel value $I_q$ also highly depends on the feature representations extracted from the LR input. In the literature, the dominant features are either extracted by convolution layers or transformer layers. Convolutions are generally high-frequency filters and the stack of convolution layers can take advantage of correlations in a large receptive field. In contrast, the aggregation transformer is a low-frequency filter and the window-based transformer constrains the correlations to be explored inside the window. Therefore, some methods propose to stack convolution and attention layers in a sequential manner \cite{lim2017edsr, zhang2018rcan, liang21swinir}  to serve as the feature extraction backbone. Unfortunately, this sequential manner can only model either local (convolution layer) or non-local (self-attention layer) information in one layer, which discards locality during global modeling or vice versa. It is well known that the local details contain rich high-frequency information while the repeated global structure is dominated by low frequencies.   
Therefore, the sequential connection used in previous SR methods cannot well represent both high-frequencies and low-frequencies at the same time. Considering that there are many sharp edges and repeat patterns in screen contents, designing a feature extraction backbone to simultaneously model high and low frequency information will be beneficial.
Therefore, we propose a novel dual branch block (DBB), which combines self-attention with depth-wise convolutions in parallel. 

This work is an extension of our previous conference work \cite{yang2021itsrn}. In this work, we make several key modifications to significantly improve the SR results while reducing the computing costs and these key modifications are also the main contributions of this work, which are summarized as follows. 
\begin{enumerate}
\item \textbf{Periodic modulated implicit transformer based upsampler}. The pixel features in continuous space are actually the modulated version of the features in the discrete space. Therefore, we propose to modulate the pixel features in discrete space via a periodic nonlinear function to generate features for continuous pixels. Then, these continuous features are transformed to pixel values via MLP. 
%
%
\item \textbf{Enhanced Transformer Based Feature Extraction}.
Considering the complementary of convolution layers and transformer layers, we propose a DBB to combine self-attention with depth-wise convolutions parallelly. Compared with the sequential connection, the parallel combination can simultaneously model high and low frequency information.  


%
\item \textbf{Large-scale dataset and SOTA performance}.
To cope with the development of transformer models,  we construct a large-scale screen image dataset, i.e., SCI2K.  It contains 2,000 SCIs with a resolution of 2K and covers various scenarios. Experiments on SCIs
and natural images demonstrate that the proposed method outperforms state-of-the-art SR methods for both continuous and discrete upsampling scales with less computing complexity. The proposed upsampler and feature extraction backbone can be directly plugged into other image SR methods to further improve their performance. 

\end{enumerate}

\section{Related Work \label{sec:related}}
    \subsection{Continuous Image Super-Resolution}
Image SR aims to recover HR images from LR observations, which is one of the most popular tasks in the computer vision community. Many deep learning based methods have been proposed for super-resolving the LR image with a fixed scale upsampler \cite{lai2017lapsrn,zhang2018rcan,lim2017edsr, shi2016subpixel,niu2020HAN,li2019SRFBN,tong2017srdensenet, Mei_2021_NLSN,zhang2019RNAN, chen2021IPT, liang21swinir,Mei2020CSNLN,xia2022ENLCA}. 
In recent years, several continuous image SR methods \cite{hu2019metasr,wang2021arbsr,chen2021liif, yang2021itsrn,lee2021lte} are proposed in order to achieve SR with arbitrary scale. The main difference between continuous SR and single-scale SR is the upsampler module. \textcolor{red}{MetaSR \cite{hu2019metasr} and ArbSR \cite{wang2021arbsr} utilize dynamic filter network as the upsampler.} 
Specifically, MetaSR \cite{hu2019metasr} introduces a meta-upscale module to generate continuous magnification.  
ArbSR~\cite{wang2021arbsr} performs SR with a plug-in conditional convolution. Inspired by implicit neural representation, some works ~\cite{chen2021liif, yang2021itsrn, lee2021lte} reformulate the SR process as an implicit neural representation (INR) problem, which achieves promising results for both in-distribution and out-of-distribution upsampling ratios. For example, LIIF~\cite{chen2021liif} replaces the meta upsampler~\cite{hu2019metasr} with MLPs, and utilizes continuous coordinates and LR features as the inputs of MLP. LTE~\cite{lee2021lte} further transforms the continuous coordinates and feature maps into 2D Fourier space and estimates dominant frequencies and corresponding Fourier coefficients for \textcolor{red}{the target value}. Different from them, our previous work~\cite{yang2021itsrn} proposes implicit transformer for upsampler and achieves promising performance on screen image SR. In this work, we further improve the implicit transformer by proposing periodic modulated implicit transformer. 
    
\subsection{SR Network Structures}
Most of deep-learning-based SR approaches focus on the feature extraction backbones after the sub-pixel convolution upsampling layer~\cite{shi2016subpixel} proposed. EDSR~\cite{lim2017edsr} builds the SR backbone with a very deep residual-skip connection structure. Motivated by the dense connection mechanism~\cite{huang2017densenet,huang2019densenet}, Tong \textit{et al.} introduce it into SR filed and proposed SRDenseNet~\cite{tong2017srdensenet}. RDN~\cite{zhang2018RDN} then further combines dense connections with residual learning to form the residual dense block (RDB). Apart from the aforementioned dense connection modules, attention modules are also widely used in SR networks. 
For example, RCAN ~\cite{zhang2018rcan} introduces SE~\cite{hu2018senet} based channel attention module to allocate more weights on important channel features and greatly improves the SR performance. Owing to the effectiveness of channel attention, spatial attention and non-local attention are also introduced to SR networks. \textcolor{red}{RNAN~\cite{zhang2019RNAN} proposes non-local attention block, where channel and spatial attentions are used simultaneously to extract hierarchical features.} Hereafter, HAN~\cite{niu2020HAN} proposes a holistic attention network, which consists of a layer attention module and a channel-spatial attention module to investigate the inter-dependencies between channels and pixels. CSNLN~\cite{Mei2020CSNLN} further presents a cross-scale non-local attention module to explore the cross-scale feature similarities and capture long-term information. Due to the success of transformer ~\cite{vaswani2017attention} in NLP and vision tasks,
it has been introduced to SR field. 
IPT~\cite{chen2021IPT} is the first that utilizes transformer for low-level vision tasks. It is pre-trained on ImageNet dataset, and then the model is finetuned on the target task, such as SR, denoising, and deraining. SwinIR~\cite{liang21swinir} adopts Swin Transformer~\cite{liu2021swin} for image restoration and has achieved outstanding performance. We observe that stacking the convolution and transformer layers sequentially (such as SwinIR) cannot model the low and high frequencies well. In this work, we propose a dual branch block to model them in parallel.      

\subsection{Implicit Neural Representation}
    Implicit Neural Representation (INR) usually refers to a continuous and differentiable function (\textit{e.g.,} MLP), which can map coordinates to a certain signal. INR is widely used in 3D shape modeling~\cite{chen2019learning,2019SAL,genova2020local,2019deepsdf}, volume rendering (\textit{i.e.,} neural radiance fields(Nerf))~\cite{2020NeRF, 2020DeRF,zhang2020nerf++, barron2021mip-nerf,barron2022mipnerf360}, and 3D reconstruction~\cite{2019occupancy, peng2020convolutional}. Inspired by INR, Chen \textit{et. al.}~\cite{chen2021liif} propose LIIF for continuous image representation, in which the image coordinates and deep features around the coordinate are transformed to RGB values. Inspired by LIIF, we propose an implicit transformer network to achieve continuous magnification while retaining the sharp edges of SCIs well.

 \subsection{Screen Content Processing}
Due to the special properties of screen contents, there are many processing tasks specifically designed for screen contents. For example, HEVC is designed for general video coding, but it is ineffective for lines, text, and graphics borders, which are the dominating objects of screen content. Therefore, HEVC-SCC \cite{liu2015hevc-scc} is proposed for screen content image compression by introducing new models, such as intra-block copy, palette mode, adaptive color transform, etc. Nowadays, screen content video (image) compression has become a classic research topic~\cite{wang2017utility,peng2016overview,tang2022tsa,wang2022perceptually} due to the explosively increasing of screen content videos (images). For image quality assessment, the screen contents are also considered separately \cite{yang2015SIQAD,gu2016learning,SCID2017,ni2016gradient,min2021screen} since the quality measurements for natural images are not suitable for screen contents. However, there is still no work exploring screen content image SR (except our conference work \cite{yang2021itsrn}), which is beneficial for the display and transmission of screen contents. In this work, we propose a parallel feature extraction module and a modulated implicit transformer to improve the screen content SR performance.   

\begin{figure}
	\centering
	\footnotesize
	\begin{tabular}{c}
		\centering
		\begin{adjustbox}{valign=c}
			\begin{tabular}{c}
             \includegraphics[width=0.44\textwidth]{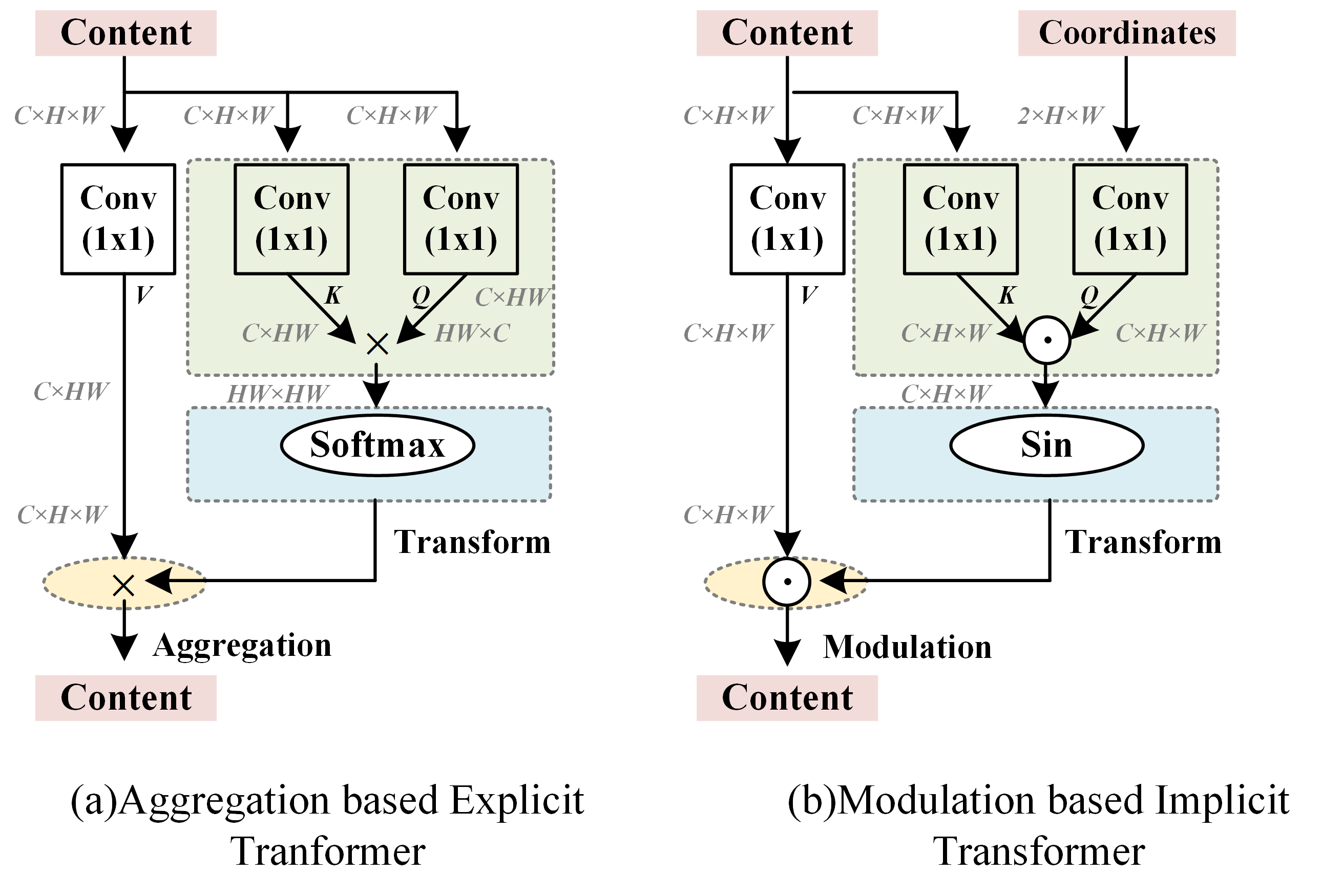}
				\\[1ex]
		
				\end{tabular}
		\end{adjustbox}

	\end{tabular}
	\caption{
	\textbf{Illustration of aggregation based explicit transformer and modulation based implicit transformer.}
}
	\label{fig:transformers}
\end{figure}



\section{Implicit and Enhanced Transformer Network for SCISR}

\label{sec:framework}
%
In this section, we first give the definition of \textit{Transformer}, and then describe the design of our implicit transformer based upsampler, followed by illustrating the proposed enhanced transformer based feature extraction backbone. 
\subsection{Transformer \label{sec:transformer}}
The key module in transformer network is the (multi-head) self attention \cite{vaswani2017attention}, which can be expressed as 
\begin{equation}
\label{eq:transformer}
z = \text{Softmax}(QK^T/\sqrt{D}+B)V,
\end{equation}
where $Q$, $K$, and $V$ denote the query, key, and value respectively. $B$ is the position encoding and $D$ is the dimension of $Q$ and $K$. $z$ is the token (i.e., $V$) aggregation result with the weights calculated based on the normalized ``distance" between $Q$ and $K$. For multi-head self-attention, this process is performed for $h$ (the head number) times parallelly and the results are concatenated along the $D$ dimension. In this work, we reformulate the transformer into a more general form, i.e., 
\begin{equation}
\label{eq:transformer-general}
z = \Phi(Q,K)\otimes V,
\end{equation}
where $\Phi(Q,K)$ denotes the weights calculated by $Q$ and $K$, and $\otimes$ represents matrix multiplication or point-wise multiplication. When $\otimes$ represents matrix multiplication and $V$ is a set of multiple token features (namely $V=[V_1, V_2,...,V_m]$), Eq. \ref{eq:transformer-general} is an aggregation transformer, where the result $z$ is the aggregation of different tokens $\{V_i\}$.  
When $\otimes$ represents point-wise multiplication ($\odot$) and $V$ is a single token feature (namely $V=V_i$), Eq. \ref{eq:transformer-general} is a modulation transformer and the result $z$ is the modulation of the current input $V_i$.  



In the literature, the predominant transformer is the aggregation transformer. Here, we denote it as explicit transformer, where $Q$, $K$, and $V$ are inferred from the same kind of inputs. For example, in SwinIR \cite{liang21swinir}, $Q$ is the linear transform of the current token (feature) while $K$ and $V$ are the linear transform of its neighboring tokens (features). In contrast, for implicit transformer, we mean $Q$ is derived from coordinates while $V$ is derived from pixel values, similar to the implicit function defined in Nerf \cite{2020NeRF}. Correspondingly, in this work, we propose the modulation transformer to module the implicit transformer. \figref{transformers} presents the summarization of the two kinds of transformers.     
\begin{figure}
	\centering
	\footnotesize
	\begin{tabular}{c}
		\centering
		\begin{adjustbox}{valign=c}
			\begin{tabular}{c}

               \includegraphics[width=0.48\textwidth]{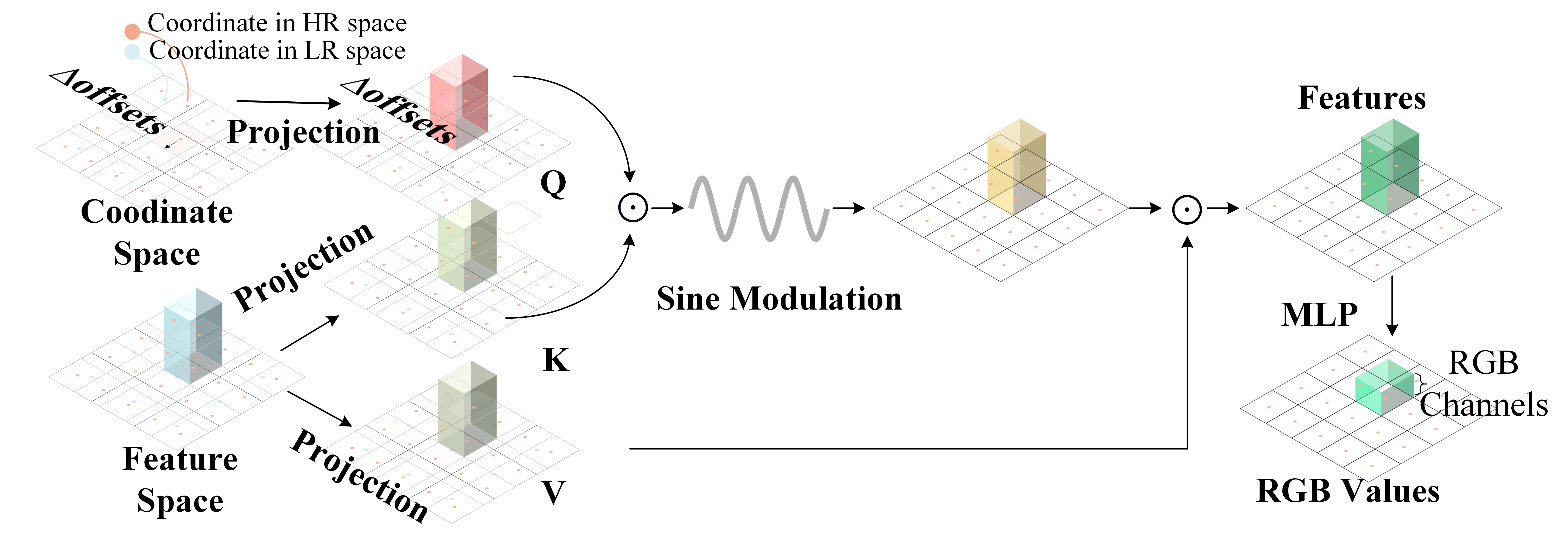}
				\\[1ex]
				\end{tabular}
		\end{adjustbox}
	\end{tabular}
	\caption{
	\textbf{The proposed implicit transformer based upsampler, which can generate pixel values in continuous space. The orange coordinates are in HR space and the blue coordinates are in LR space.}
	}
	\label{fig:upsampling}
\end{figure}
\begin{figure*}[ht]
	\centering
	\footnotesize
	\begin{tabular}{c}
		\begin{adjustbox}{valign=c}
			\begin{tabular}{c}
				\includegraphics[width=1\textwidth]{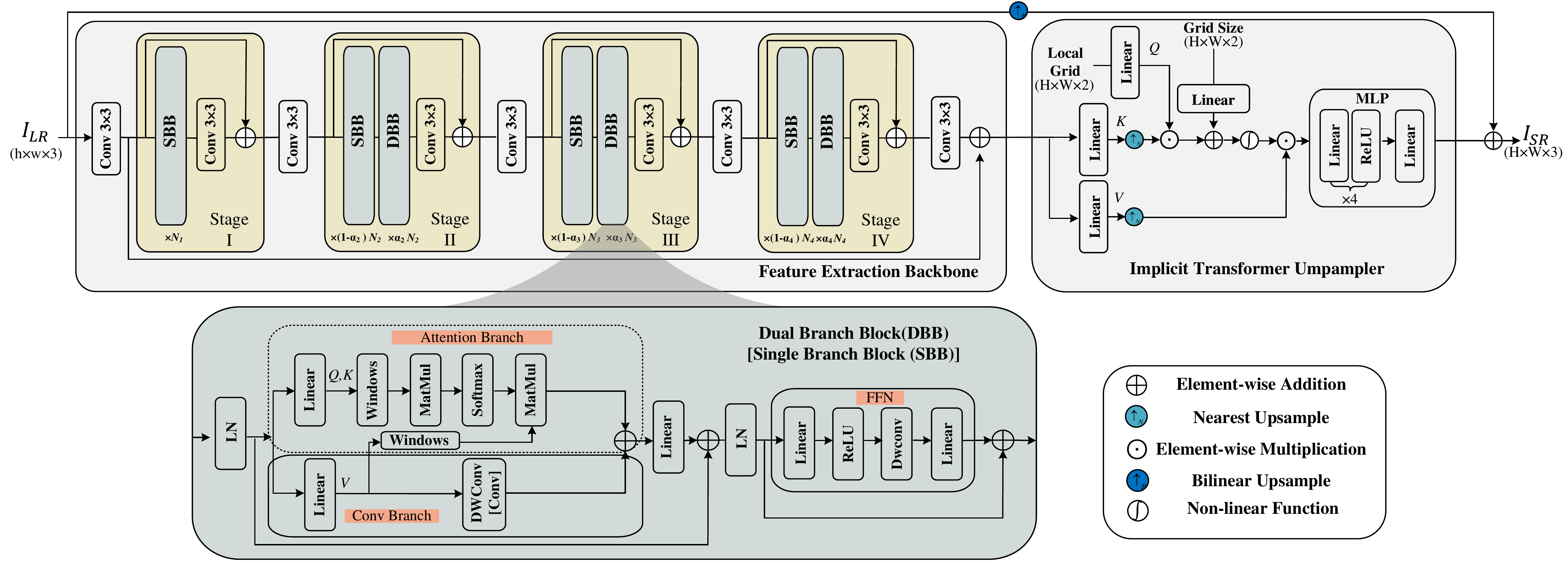}
				\\[1ex]
			\end{tabular}
		\end{adjustbox}
	\end{tabular}
	\caption{
	\textbf{Detailed network architecture of the proposed ITSRN++}. Our model includes two parts: enhanced transformer based feature extraction backbone and implicit transformer based upsampler. The feature extraction backbone contains four stages, with dual branch block and single branch block. The DBB is constructed by attention branch and conv branch. Removing the attention branch (circled by dotted line) and changing the conv block to dwconv block resulting in SBB.}
	\label{fig:net_arch}
\end{figure*}
\subsection{Implicit Transformer Based Upsampler} 
Before introducing our upsampling scheme, we first revisit the image upsampling process. If we view image upsampling in a continuous space, it can be seen as sampling values at discrete points within a finite area. For image interpolation, suppose that we have an LR image $I_{LR}$ that needs to be upsampled. The pixel value of query point $q(i,j)$ in the HR image $I_{HR}$ is obtained by fusing pixel values of its neighboring key points $k(\Omega (i), \Omega (j))$ in $I_{LR}$ with weighting coefficients, where $\Omega (i)$ denotes the neighboring points of $i$ in $I_{LR}$. Denoting the \textit{query} points in upsampled image as $Q$, the \textit{key} points in the input LR image as $K$, and the \textit{value} on the corresponding key points as $V$, the image upsampler can be reformulated as Eq. \ref{eq:transformer-general}, \textit{i.e.}, an aggregation transformer~\cite{vaswani2017attention}.
Instead of utilizing pixel features to generate $Q$ and $V$, the interpolation transformer deals with pixels' coordinates and their values. \textcolor{red}{Inspired by the implicit function in NeRF~\cite{2020NeRF}, which utilizes the pixel coordinates to generate RGB values, we rename the interpolation process as \emph{Implicit Transformer}, and propose a novel Implicit Transformer Network for SCI SR.}

Specifically, we redesign the upsampling process as   
\begin{gather}
    \textit{Offsets} = P_{HR} - P^*_{LR}, \quad Q = \text{Linear}(\textit{Offsets}), \notag\\
    K = \text{Linear}(\mathcal{F}_{LR}), \quad V =\text{Linear}(\mathcal{F}_{LR}),
\label{eq:QKV}
    \end{gather}
where $\textit{Offsets}\in \mathbb{R}^{H\times W\times 2}$ denotes the relative offsets between the query points in the HR space ($P_{HR}$) and their corresponding nearest neighboring points ($P^*_{LR}$) in the LR space, as shown in Fig. \ref{fig:upsampling} (the orange pixel in the HR space has a nearest neighbor in the LR space denoted by blue pixel). To ease the training process, we normalize the coordinates into the range of [-1, 1]. Different from \cite{chen2021liif,yang2021itsrn, lee2021lte}, which directly predict (using the function $\Phi(\cdot)$) the continuous representation with the relative coordinate $p_q-p^*_k$, we treat it as $\textit{query}$ to  \textcolor{red}{perform local continuous representation}. 
%
%
%
We project the $\textit{Offsets}$ to $Q\in \mathbb{R}^{H\times W\times C}$ with a linear transform, and project the deep features (denoted as $\mathcal{F}_{LR}$) extracted from the LR input to $K,V \in \mathbb{R}^{H\times W\times C}$ with another two linear transforms. For the query point $p$ (in the continuous HR space), the query feature is $Q_p$, and its corresponding key and value in the LR space is $K_{p^*}$ and $V_{p^*}$, where ${{p^*}}$ represents the nearest neighboring point of $p$. We observe that utilizing the aggregation of different ${V_i}$ to predict pixel value may lead to smoothness due to the low-pass characteristics of the weighted aggregation process. Therefore, we propose the modulation transformer as follows:  
\begin{equation}
I_{p} = \Phi(\sigma(K_{ {p^*}}\odot Q_p + \text{Linear}(S))\odot V_{p^*}),  
\label{Eq:point-transformer}
\end{equation}
where $\sigma$ is the nonlinear mapping function, and $\Phi$ represents an MLP with four layers.
The element-wise multiplication between $K_{p^*}$ and $Q_p$ generates the attention weight $\in \mathbb{R}^{1\times 1\times C}$. Similar to \cite{liu2021swin}, we further introduce a scale bias $S\in \mathbb{R}^{1\times 1\times 2}$, which refers the scaling factor along $H$ and $W$ dimension, respectively. We project $S$ with a linear layer to make it have the same dimension ($1\times1\times C$) as that of the weights. To reweight the attention matrix, many non-linear mapping functions can be adopted, such as \textit{Softmax}, \textit{Sigmoid}, and \textit{Tanh}. However, these are all monotone increasing functions, which may limit the non-linear mapping space. 
%
%
Some works \cite{sitzmann2020SIERN,tancik2020fourier} demonstrate that the periodic activation functions can facilitate networks to learn  high-frequency information. Therefore, we propose to utilize a periodic function to reweight the attention values. 
In this work, we utilize $sin(\cdot)$  as the nonlinear mapping function to reweight the weights. Hereafter, we modulate the value $V$ with the recalibrated weights via point-wise multiplication. In other words, $V_p$ is a modulated version of $V_{p^*}$.
\textcolor{red}{The features in discrete space can be modulated to features in continuous space by the learned non-linear weights.}
Finally, we use an MLP to map the feature $V_p$ to pixel value $I_p$. The whole process is illustrated in Fig. \ref{fig:upsampling}. 


Utilizing Eq. \ref{Eq:point-transformer} to generate the pixel value can get rid of the influence of its neighboring pixels. This is beneficial for high-frequency detail generation but it may lead to discontinuity. Therefore, we further propose to utilize a local filter to refine the results. This process is denoted as:
\begin{equation}
    \hat{I}_{q} = \sum_{p \in \Omega_(q)} \omega(p,q)I_q,
\end{equation}
where $\hat{I}_{q}$ is the refined pixel value. $\Omega (q)$ is a local window centered at $q$, and $w$ represents the neighbors' (denotes by $p$) contribution to the target pixel. In this work, we directly utilize the bilinear filter as the weighting parameter $w$. 

We would like to point out that our upsampler in this work is totally different from LIIF and ITSRN. 
\textbf{1) Relationship with LIIF.}
In implicit function based SR method LIIF \cite{chen2021liif}, $K$ is the nearest neighbor coordinate in $I^L$ for the corresponding $Q$ in $I^H$, and $V$ is the corresponding deep features of $K$. Different from ours, the super-resolved pixel value $I_q$ is obtained by concatenating $V$ and the relative offsets between $Q$ and $K$ first and then going through the nonlinear mapping $\Phi$ (realized by an MLP). It has achieved promising results due to the strong fitting ability of the MLP. However, we observe that directly concatenating the extracted pixel features and the relative coordinates is not optimal and this will lead to a large resolution space. Therefore, we utilize $\Phi(Q, K)V$ other than $\Phi(Q, K, V)$ for upsampling. In this way, we constrain the features in continuous space to be a variant of its neighboring feature in the discrete space and the pixel value is inferred from the modulated features.   
\textbf{2) Relationship with ITSRN.} In our previous work, we model the upsampler as 
\begin{equation}
    I = \Phi(Q,K)V = H(Q-K)V,
\end{equation}
where $H(Q-K)$ maps the relative offsets between the HR coordinates and the corresponding LR coordinates to high-dimensional weights, which are then multiplied with the deep pixel features $V$ to generate the pixel value $I$. However, it consumes lots of memory and $H(Q-K)$ cannot be adaptive according to the image contents. In contrast, our upsampling weights in  Eq. \ref{Eq:point-transformer} can be adaptive according to the image contents.  

\subsection{Enhanced Transformer Based Feature Extraction}

\label{sec:architecture}
%
As shown in Eq. \ref{eq:QKV} and \ref{Eq:point-transformer}, the upsampling results heavily depend on the extracted features $\mathcal{F}_{LR}$. In this work, we propose an enhanced transformer based feature extraction backbone, which is constructed by cascaded dual-branch block (DBB) and single-branch block (SBB), as shown in \figref{net_arch}. In the following, we first give an overview of the network structure and then present the details of the proposed DBB and SBB.   



\textbf{Network Overview.} 
Our feature extraction backbone is constructed based on Swin Transformer \cite{liu2021swin}, and we incorporate several key modifications to model the sharp edges of screen content images. Following \cite{liang21swinir}, for the input low-resolution image $I_{LR} \in \mathbb{R}^{ H \times W \times C_{in} }$, where $H$, $W$, and $C_{in}$ are the  height, width, and channels of the input image, respectively, we utilize one $3\times 3$ convolution layer to extract its shallow feature $\mathcal{F}_s \in \mathbb{R}^{H \times W \times C}$. Then $\mathcal{F}_{s}$ goes through the deep feature extraction module to generate the final features 
$\mathcal{F}_{LR}$ for the LR input. The deep feature extraction module contains four stages, and each stage contains $(1-\alpha)N\times$ SBB, $\alpha N \times$ DBB, and one $3\times 3$ convolution layer.

\textbf{Dual Branch Block.}
As mentioned in \secref{transformer}, the classical transformer block, which is constructed by multi-head-self-attention (MHSA) and feed-forward network (FFN) tends to generate smoothed results since the fusion coefficients generated by \textit{Softmax} function are all positive values. Therefore, we propose to introduce a convolution branch to enrich the feature representations of the transformer block, constructing the Dual Branch (attention branch and conv branch) Block:
\begin{equation}
    \text{DBB}(\mathcal{F})= \text{Attention}(\mathcal{F}) + \text{Conv}(\mathcal{F}),
    \label{eq:dbb}
\end{equation}
where $\mathcal{F}$ is the input feature map $\in \mathbb{R}^{H\times W \times C}$.
Specifically, our attention branch is similar to that in \cite{liang21swinir}\cite{liu2021swin}. At each stage, the input feature map $\mathcal{F}$ is projected to $Q, K, V$ with linear transform layers, where $H\times W$ is the token number (\textit{i.e.}, a point is a token) and $C$ is the channel number. These tokens are further partitioned into non-overlapped $M\times M$ windows, and the window number is $HW/M^2$. Therefore, in each window, there are $M^2$ tokens and the dimension of each token is $D$. Then, the self-attention is calculated within the local window, namely
\begin{equation}
    \text{Attention}(Q,K,V)= \text{Softmax}(QK^T/\sqrt{D}+B)V,
	\label{eq:attention}
\end{equation}
where $B\in \mathbb{R}^{M^2\times M^2}$ is the learnable relative position bias. To reduce the computing complexity, following \cite{liu2021swin}, we perform the attention process parallelly for $h$ times and they are concatenated along the $D$ dimension, constructing MHSA.
We denote the features after the MHSA as $\mathcal{F}_{\text{MHSA}}$. Since the windows are not overlapped, there are no connections among different windows. Therefore, following the shifting window strategy proposed in  \cite{liu2021swin} to merge information across local windows, we stack regular and shifted partitions sequentially along the layers. Note that, the shifted window partitioning denotes shifting the tokens by half size of the windows. 

For the convolution branch, different from ~\cite{chen2022mixformer,si2022inceptionformer,chen2022HAT}, which perform convolution with the input features, we instead extract the convolution features from the value $V$, which is not partitioned into windows. In this way, the convolution layer can explore the correlations among neighboring windows, which further enhances the correlations of tokens along the window borders. In DBB, We use DWConvBlock (DWConv(k=5)-Relu-Conv(k=1)-CA) as Conv(). After going through the depth-wise convolution, point-wise convolution, and channel attention layers, we obtain the conv features $\mathcal{F}_{\text{conv}}$. Then, $\mathcal{F}_{\text{conv}}$ and  $\mathcal{F}_{\text{MHSA}}$ are added together, going through another linear layer. Hereafter, a multi-layer perceptron (MLP) that has two fully connected layers with ReLU non-linearity between them is used as FFN for further feature aggregation. Besides, a pre LayerNorm (LN) layer is inserted before FFN and MHSA+Conv modules, and the residual skip connection is added for both blocks, as shown in \figref{net_arch}.  

\textbf{Single Branch Block.} Note that, DBB consumes lots of memory and computing resources. \textcolor{red}{To reduce the computing complexity, we also construct the single branch block (SBB), where the attention branch (circled by a dotted line in \figref{net_arch}) is removed. In this case, We replace the DWConvBlock with ConvBlock(Conv(k=3)-ReLU-Conv(k=3)-CA). For earlier stages, we utilize more SBBs to extract visual features while in later stages we utilize more DBBs to enhance the feature aggregation process. The ratios ($\alpha_2$,$\alpha_3$,$\alpha_4$) between the number of DBB and SBB are explained in the experiments.}

\begin{figure}
	\setlength{\fs}{-0.4cm}
	\scriptsize
	\centering

	\begin{tabular}{cc}
	 Convolution-Branch  & Attention-Branch 
	 \\
	 \hspace{\fs}
	 {\includegraphics[width=0.23\textwidth]{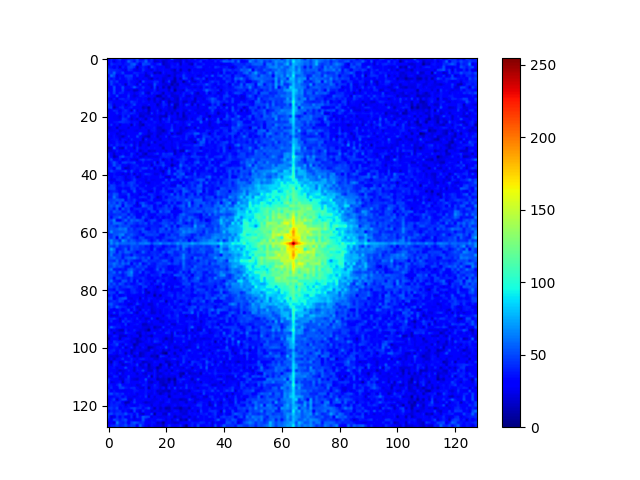} \hspace{\fs}}
	 & {\includegraphics[width=0.23\textwidth]{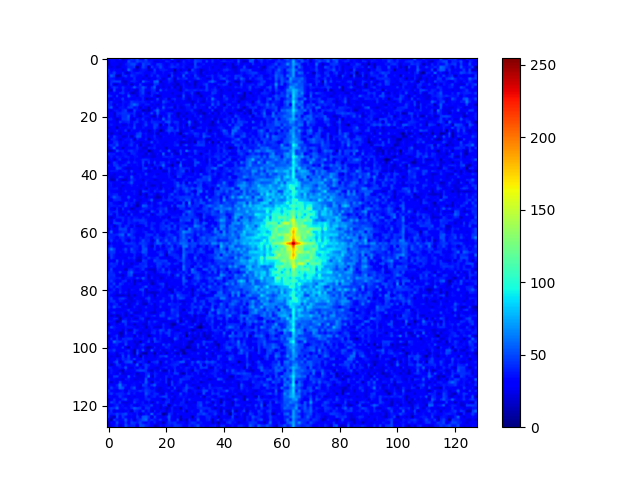} \hspace{\fs}} 
	 
	
	\\
	\end{tabular}

	\caption{
		\textbf{Centralized Fourier spectrum of conv-branch feature and attention-branch feature in the second DBB. Note that, the brighter the color, the greater the amplitude of the Fourier spectrum.}
	}
	\label{fig:freq}
\end{figure}

\figref{freq} presents the frequency statistics of the features learned in convolution branch and attention branch. Take the features in the second DBB for example, we perform Fourier transform on the features $\mathcal{F}_{\text{conv}}$ and  $\mathcal{F}_{\text{MHSA}}$, and the features along different channels are averaged. It can be observed that the features learned by the conv-branch contain more high-frequency information compared with those learned by the attention-branch. In this way, the conv-branch feature is a good complementary of the attention-branch feature.    

\subsection{Multi-scale Training}
\label{sec:multiscale}
To enable multi-scale training, we construct the mini-batch with upsamping scales uniformly sampled from $\mathcal{U}(1,4)$. Let $r$ denote the scaling factor, and  $h, w$ are the height and width of the LR patch. We first crop the HR patch with a size of $rh \times rw$ from the HR image to serve as ground truth (GT). Then we generate its LR counterpart by downsampling the GT with the scale factor $r$ via bicubic interpolation. Finally, we randomly sample $hw$ pixels from each GT patch to make the GTs in a batch have the same size. Meanwhile, for a batch of LR inputs, they share the same size but have different magnification ratios. Compared with training with a single magnification factor for each batch, multi-scale training is more stable and leads to better results. 

During training, we utilize the most popular $\ell_{1}$ loss function, namely 
\begin{equation}
    \mathcal{L} = \left \| I_{SR} - I_{HR} \right \|_1,
\end{equation}
where $I_{SR}$ is the SR result, and $I_{HR}$ is the GT image.



\section{SCI2K dataset}
\begin{figure}
	\centering
	\footnotesize
	\begin{tabular}{c}
		\centering
		\begin{adjustbox}{valign=c}
			\begin{tabular}{c}
				\includegraphics[width=0.45\textwidth]{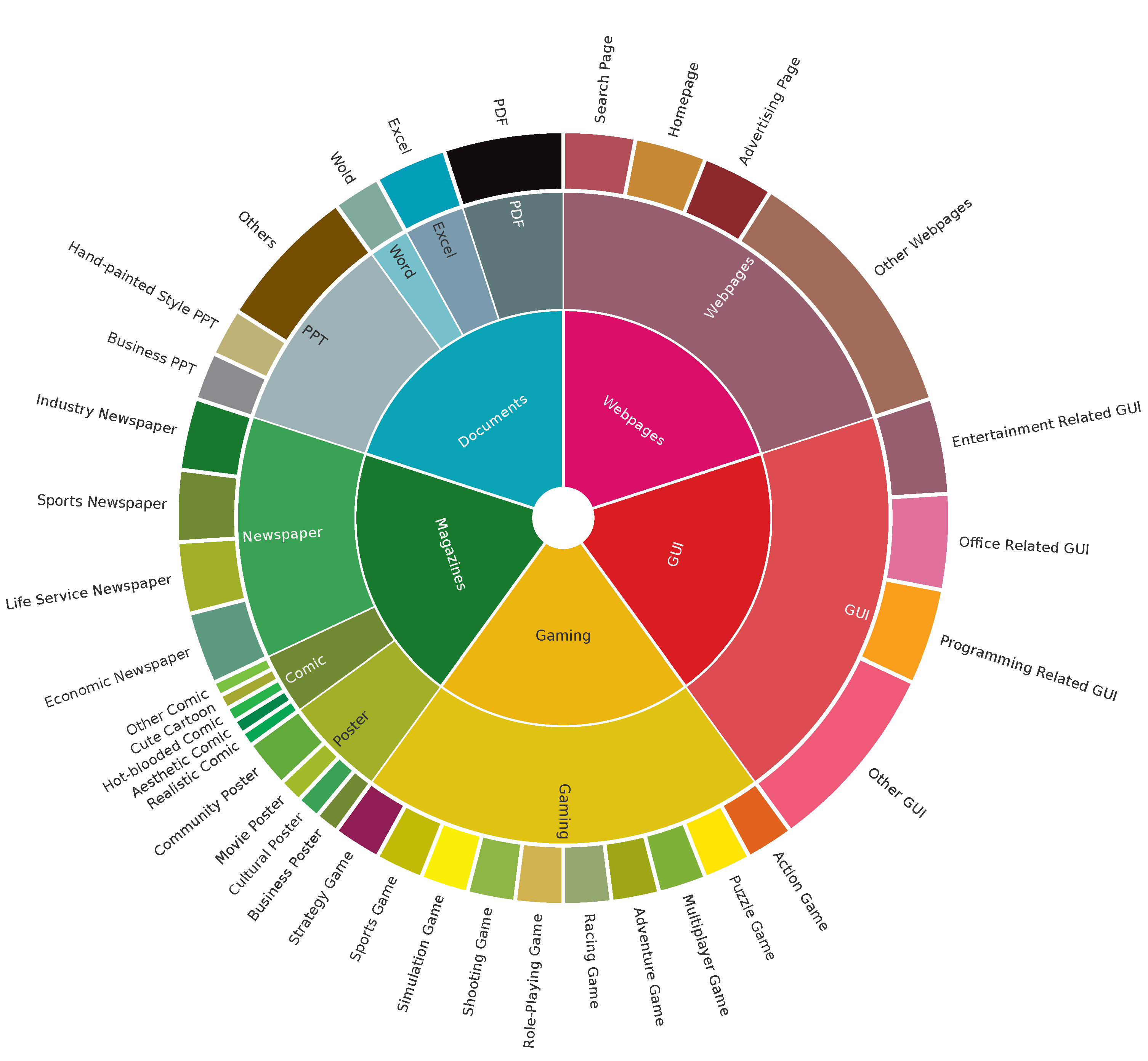}
			\end{tabular}
		\end{adjustbox}
	\end{tabular}
	\caption{
	\textbf{The radial dendrogram of our SCI2K dataset, which contains five main categories of screen contents.}
	}
	\label{fig:dataset}
\end{figure}

Compared with the datasets for natural image SR, such as DIV2K~\cite{agustsson2017div2k}, Urban100~\cite{huang2015urban100}, and BSDS100~\cite{martin2001b100}, datasets for SCIs are limited and most of them are for quality assessment. For example, the SIQAD dataset~\cite{yang2015SIQAD}, designed for image quality assessment, contains 20 ground truth images (with resolution around $600\times800$) and 980 distorted images. The SCID \cite{SCID2017} dataset consists of 40 screen content images (with a resolution of $1280\times720$) and 1800 distorted images. The CCT dataset \cite{CCT} consists of three kinds of image contents, i.e.,  natural scene image, computer graphic image, and screen content image, with 24 reference images for each type. However, the image resolutions in the three datasets are mostly less than 1K and the image amount for high-quality reference images is far from enough for training SR networks. Our previous work ITSRN \cite{yang2021itsrn} constructs the first SCI SR dataset, named as SCI1K, which contains 1000 screenshots with a resolution of $1280\times720$ and $2560\times1440$ (less than 100 images).



In recent years, there is a trend to train larger models with larger datasets, and the image resolution is also increased. For example, for natural image SR, the DIV2K dataset \cite{agustsson2017div2k}, which contains 1000 images with a resolution of 2K is widely used. Some transformer-based works \cite{liang21swinir} further utilize the combination of DIV2K and Flickr2K datasets to boost the SR results. In contrast, there is still no large-scale SCI dataset. On the hand, to cope with the development of large screens, an SCI SR dataset with a larger resolution is demanded. Therefore, in this work, we construct an SCI2K dataset, which contains 2000 images with 2K resolution, by manually capturing the screen content with the snipping tool in Windows 10. \figref{dataset} lists the detailed categories of our dataset, which covers various contents, including documents, magazines, webpages, game scenes, \textit{etc}, which are common cases in our daily life. 

The 2000 images are randomly split into train and test sets, which contain 1800 and 200 images, respectively. To be consistent with previous works, the LR images are synthesized by bicubic downsampling. To further simulate the compression degradations caused by transmission and storage, we further construct the SCI2K-compression dataset by applying JPEG compression on the LR images. The quality factors are randomly selected from 75, 85, and 95.





\section{Experiments}
\subsection{Implementation and Training Details}
\textbf{Datasets.}
For SCI SR, we use our proposed SCI2K dataset for training. For evaluation, besides the testing set in SCI2K, we further utilize three benchmark screen datasets \cite{CCT,SCID2017,yang2015SIQAD}. Since these datasets are not designed for the SR task, we downsample their GTs with bicubic interpolation (realized by the imresize function in PIL package) to construct LR-HR pairs.

\textbf{Network Configurations.}
For the attention branch, the channel number for the shallow convolution layer and the last convolution layer is 64. The block numbers $N$ in the four stages are 2, 8, 8, and 16 respectively. And the channel number of four stages are 64, 64, 128 and 192, respectively. The ratios between the numbers of DBB and SBB in Stage 2-4 are $\alpha_2=0.25$,$\alpha_3=0.25$, and $\alpha_4=0.75$.  The head number in MHSA is 8. The window size $M$ is 16. 

\textbf{Training Details.}
During the training process, the batch size is set to 16 and the LR patch size is $48 \times 48$. Each epoch contains about $2,000$ iterations and the max epoch is set to 1000. Adam algorithm is used for optimization. The learning rate is initialized as 2e-4 and decayed by a factor 0.5 at the $400^{\text{th}}, 800^{\text{th}}, 900^{\text{th}}, 950^{\text{th}}$ epoch. The training data is augmented with rotation and flip with a probability of $0.5$.

\subsection{Ablation Study}
\label{sec:analysis}

In this section, we perform ablation study to demonstrate the effectiveness of the proposed implicit transformer based upsampler and enhanced transformer based feature extraction backbone. The performances of different variants are evaluated on the testing set of the proposed SCI2K dataset and SCID dataset. 

\subsubsection{Ablation on Implicit Transformer}
\begin{table}[h]
    \centering
    \caption{\textbf{Ablation study on the proposed modulation-based implicit transformer by replacing it with two variants}. The results are evaluated on SCI2K and SCID datasets for $\times 4$ SR.}
    \vspace{-3mm}
    \label{tab:ablition_on_qkv}
    \resizebox{0.48\textwidth}{!}
    {
    \begin{tabular}{l|c|c}
    \toprule
    \multirow{2}{*}{Variants}
     &  \textsc{SCI2K}  &
     \textsc{SCID}  \\
     & PSNR$\uparrow$ / SSIM$\uparrow$  & PSNR$\uparrow$ / SSIM$\uparrow$ 
    \\
    \hline
    only V (Bilinear)
     & 28.03  /  0.9275  & 24.35 / 0.8498 \\
    concatenation (LIIF's)
      & 31.25 /  0.9571 & 27.37 / 0.9073  \\

    modulation (Ours)
     & \textbf{31.66} / \textbf{0.9597} & \textbf{27.84} / \textbf{0.9149} \\
    \bottomrule
    \end{tabular}
    }

    \label{table-upsampler}
\end{table}

\begin{table}[h]
    \centering
    \caption{
		\textbf{Ablation study on the reweighting function in implicit transformer}. The results are evaluated on SCI2K and SCID datasets for $\times4$ SR.
}
    \vspace{-3mm}
    \label{tab:reweigh_func}
    \resizebox{0.48\textwidth}{!}
    {
    \begin{tabular}{l|c|c}
    \toprule
    \multirow{2}{*}{}
  
     &  \textsc{SCI2K}  &
     \textsc{SCID}  \\
     & PSNR$\uparrow$ / SSIM$\uparrow$  & PSNR$\uparrow$ / SSIM$\uparrow$ 
    \\
    
    \hline

    sin & \textbf{31.66} / \textbf{0.9597} & \textbf{27.84} / \textbf{0.9149} \\
    sin $\rightarrow$ tanh & 31.42 / 0.9587 & 27.67 / 0.9135  \\
    sin $\rightarrow$ sigmoid & 31.36 / 0.9585  & 27.58 / 0.9120 \\
    sin $\rightarrow$ softmax & 31.29 / 0.9578 & 27.52 / 0.9113  \\
    \bottomrule
    \end{tabular}
    }
\end{table}

\begin{figure*}[t]
	\setlength{\fs}{-0.4cm}
	\scriptsize
	\centering

	\begin{tabular}{cccccc}
	 HR Image & & Conv only  &  Attention only & Sequential & Parallel(Ours)
	 \\
	 \hspace{\fs}
     \includegraphics[width=0.18\textwidth]{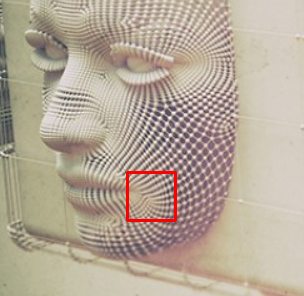}
     & \rotatebox{90}{LAM Attribution} 
	 &  {\includegraphics[width=0.18\textwidth]{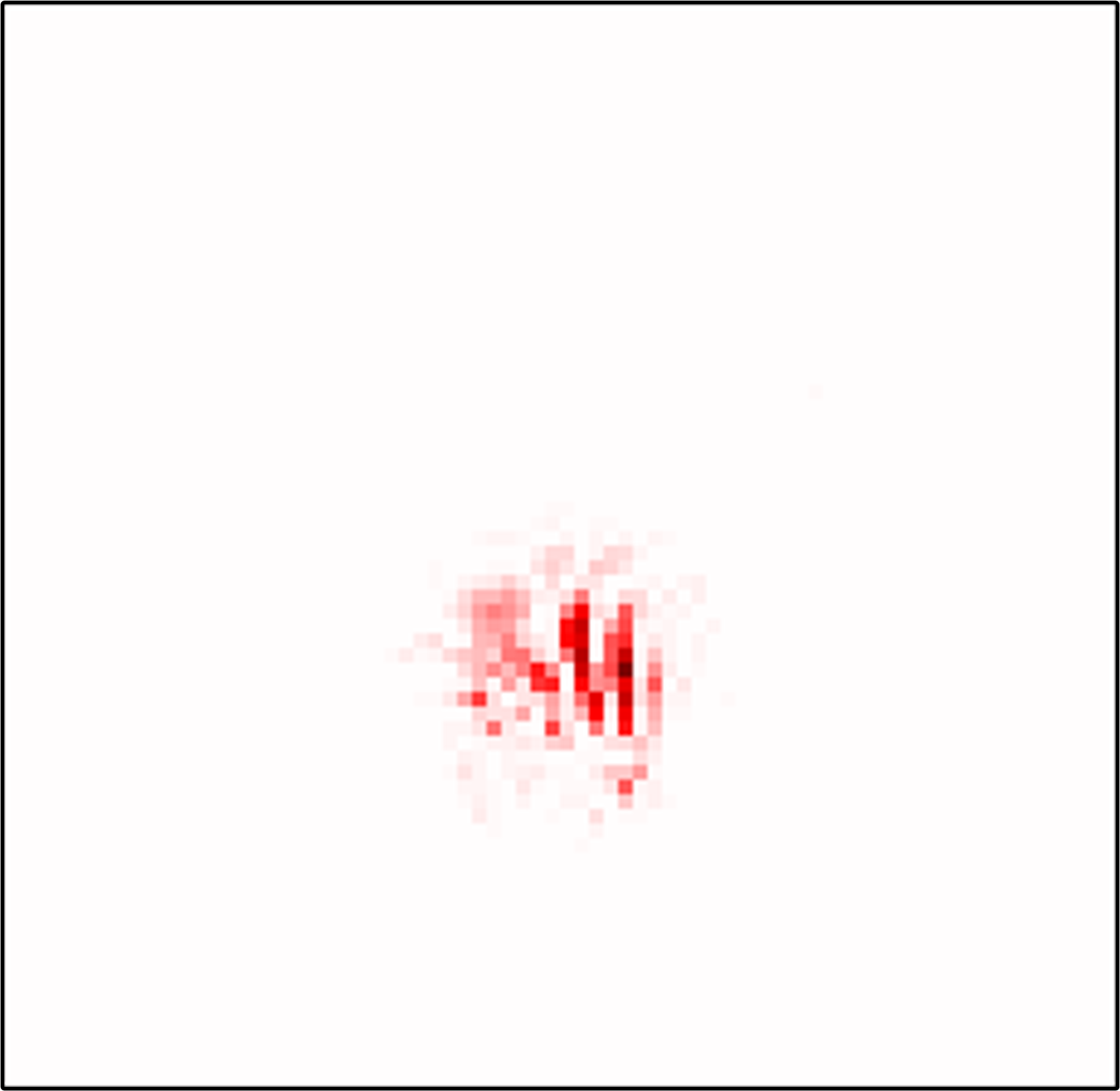} \hspace{\fs}}
	 &  {\includegraphics[width=0.18\textwidth]{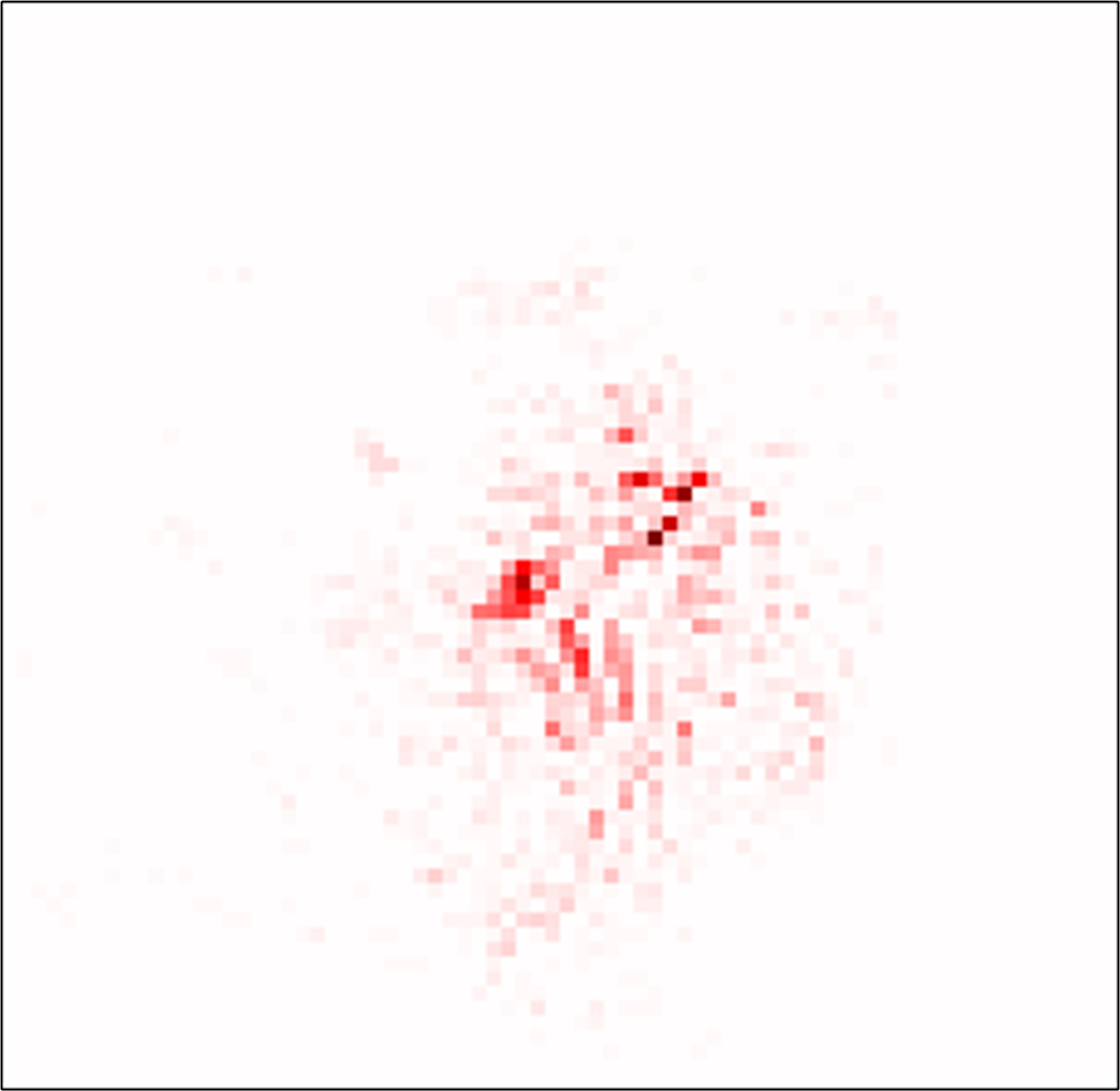} \hspace{\fs}}
	 &  {\includegraphics[width=0.18\textwidth]{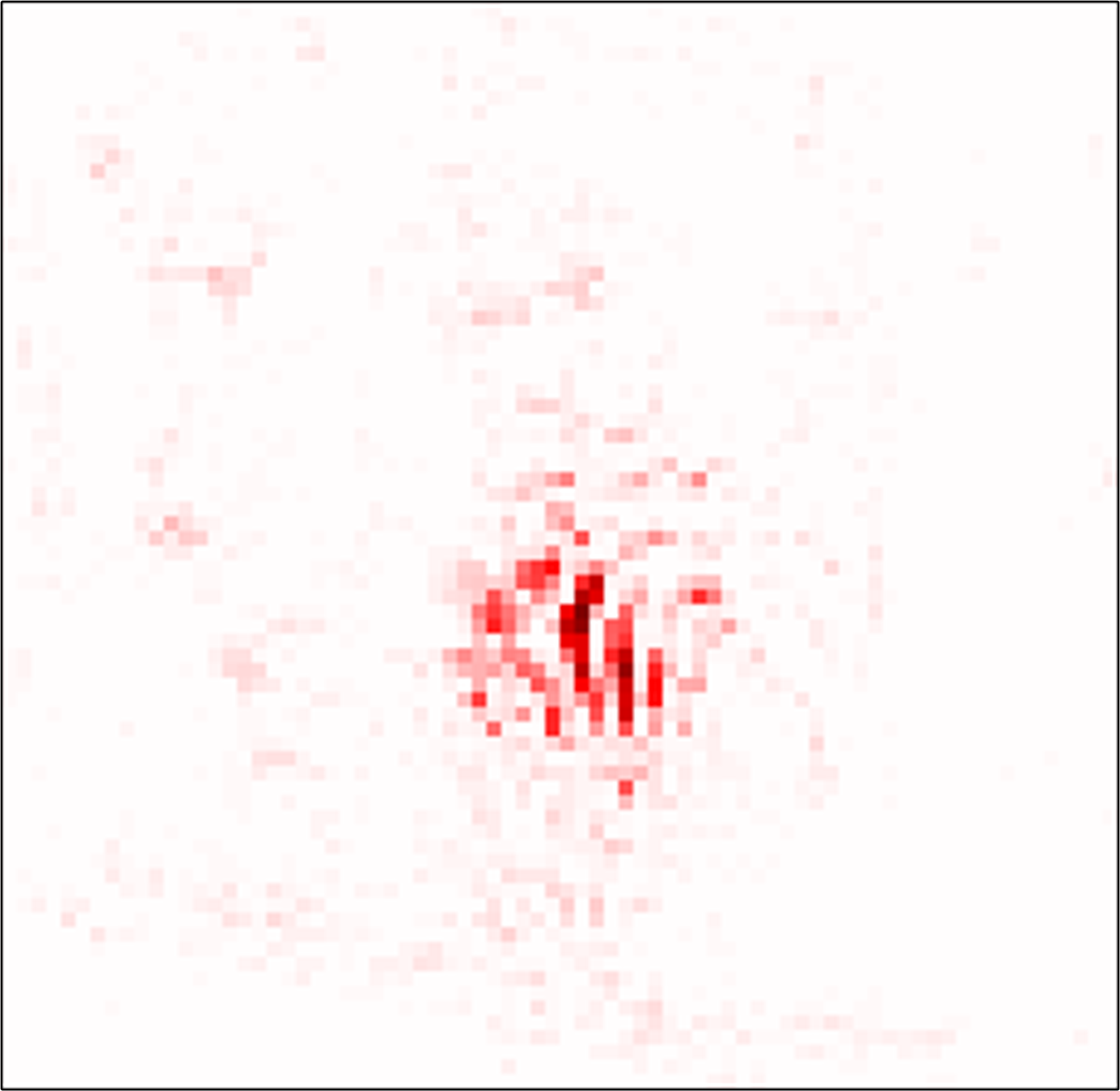} \hspace{\fs}}
	 & {\includegraphics[width=0.18\textwidth]{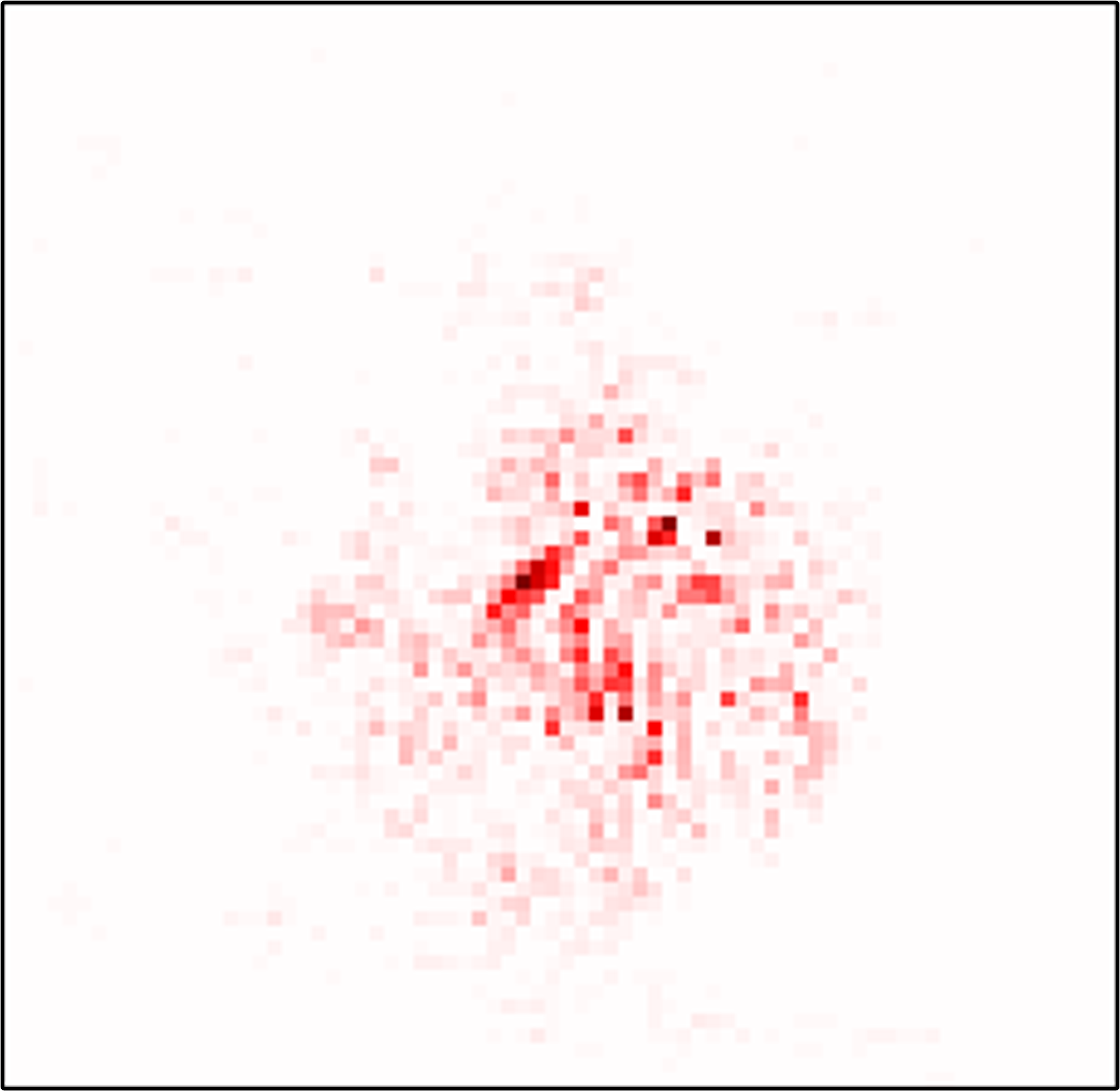} \hspace{\fs}} 
	
	\\
	 LR Image & & 5.42 & 12.09 & 18.48 & 14.07
	 
	\\
	\hspace{\fs}
	\includegraphics[width=0.18\textwidth]{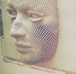}
	& \rotatebox{90}{SR Results}
	& {\includegraphics[width=0.18\textwidth]{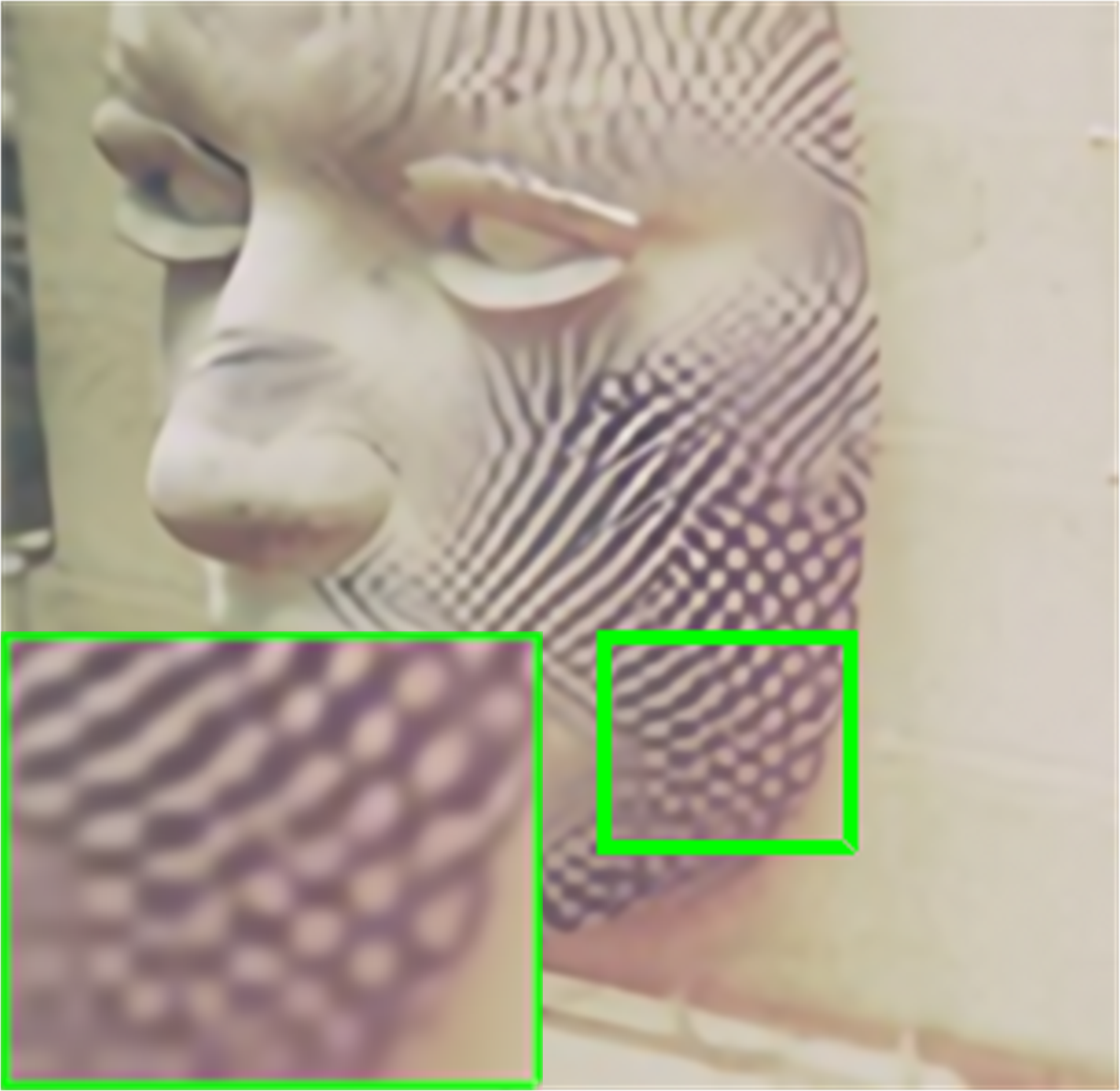} \hspace{\fs}}
	& {\includegraphics[width=0.18\textwidth]{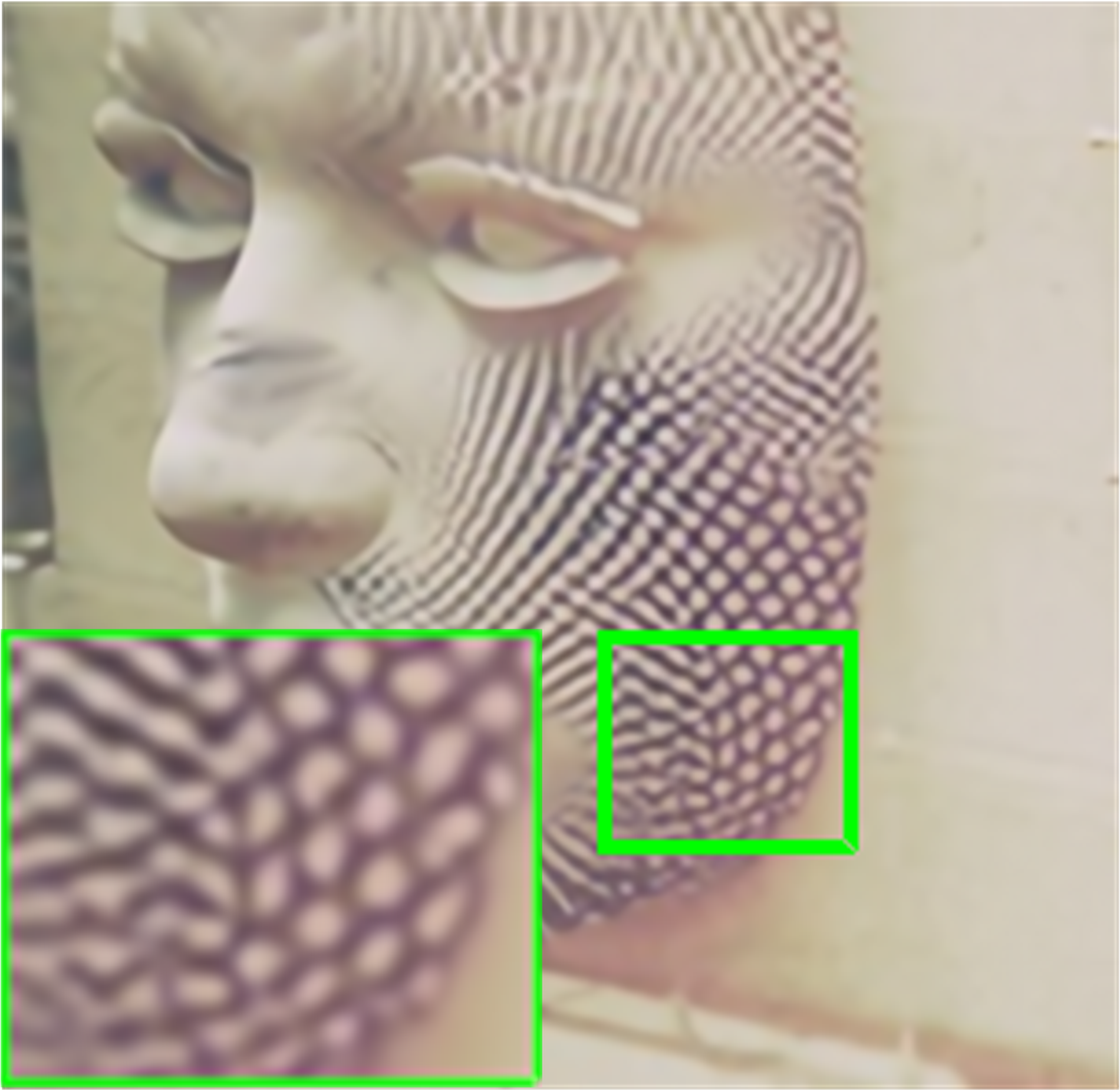} \hspace{\fs}}
	& {\includegraphics[width=0.18\textwidth]{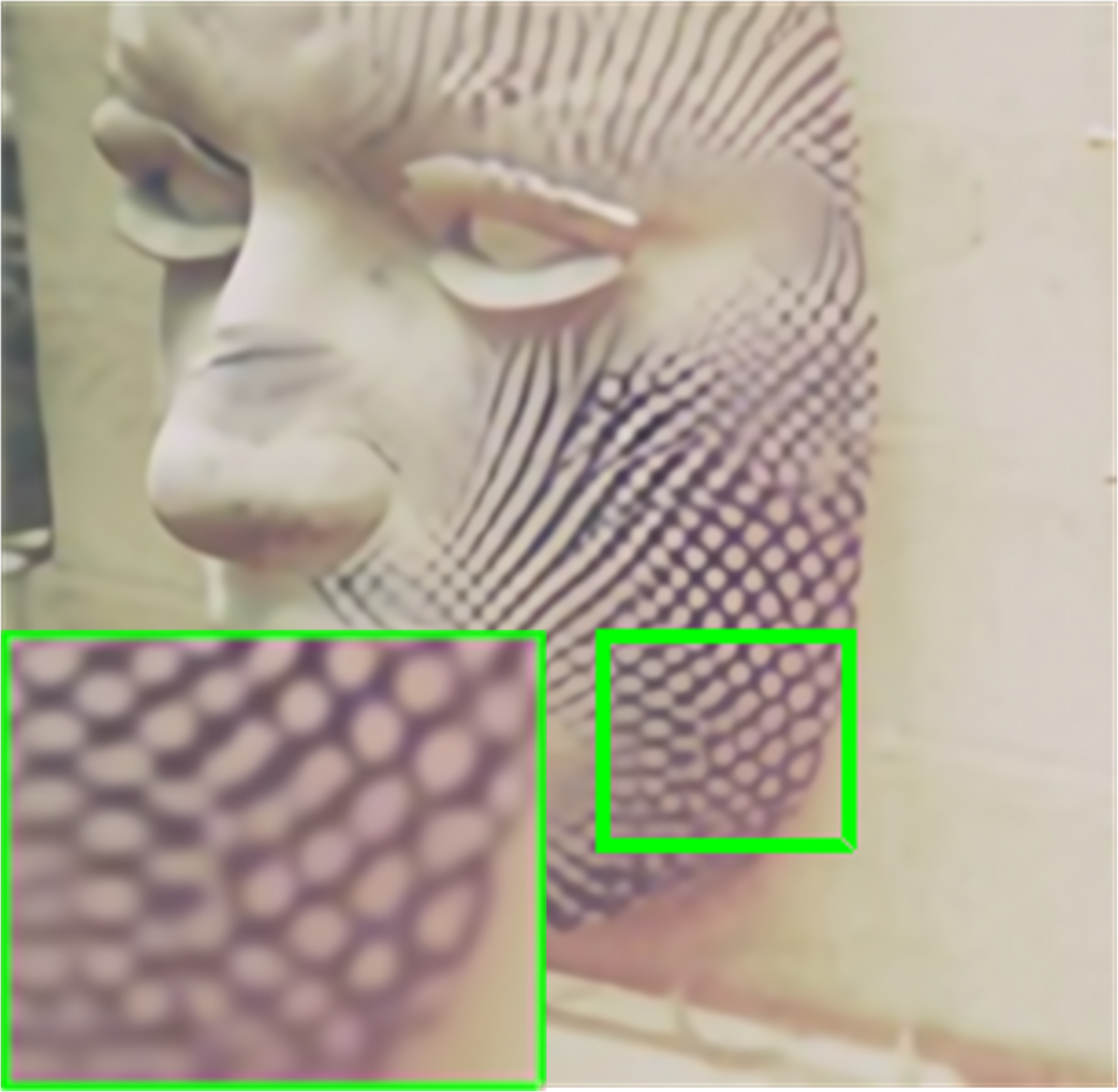} \hspace{\fs}}
	& {\includegraphics[width=0.18\textwidth]{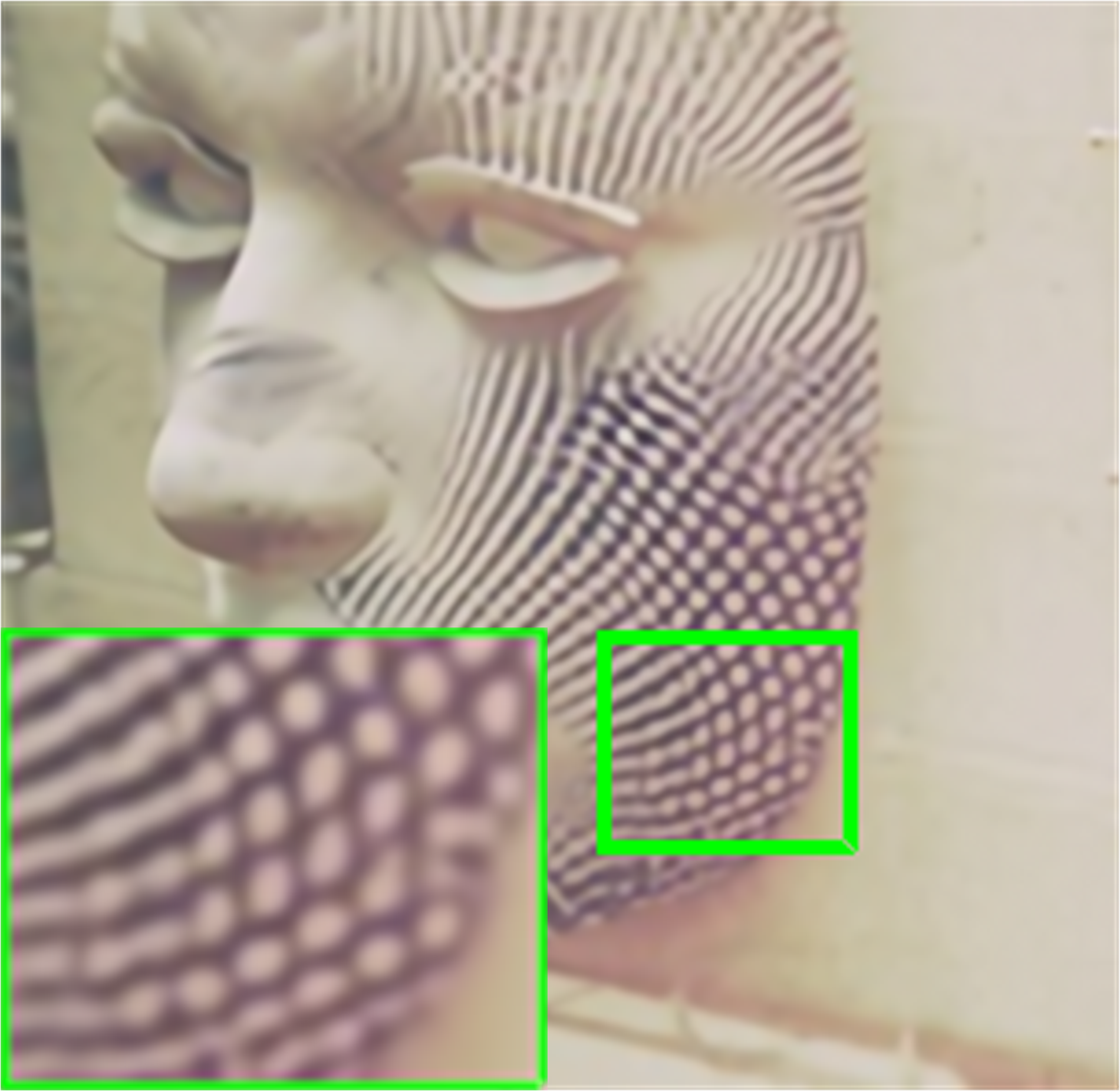} \hspace{\fs}}
	\\
	PSNR / SSIM  &   &  25.96dB / 0.7397 & 26.32dB / 0.7691 & 26.23dB / 0.7521  & 27.12dB / 0.7920
	\\

	\end{tabular}

	\caption{
		\textbf{Visual comparisons for the proposed DBB and its variants}. The first row is LAM \cite{gu202LAM}, which highlights the pixels that contribute to the reconstruction of the patch circled by the red box. DI~\cite{gu202LAM}(diffusion index) values (larger values indicate more pixels are involved) for each LAM are also given. The second row is the corresponding SR results. The PSRN and SSIM results are   
	}
	\label{fig:ablation_dpfb}
\end{figure*}

We first perform ablation on the proposed implicit transformer based upsampler (Eq. \ref{Eq:point-transformer}) by replacing it with two variants. The first variant is the upsampler in LIIF, namely 
\begin{equation}
I_{p} = \Phi(\text{concat}(Q_p, V_{p^*})),
\label{Eq:LIIF-upsampler}
\end{equation}
where $\Phi(\cdot)$ is the MLP, the same as that in Eq. \ref{Eq:point-transformer}. Note that following \cite{chen2021liif}, we utilize feature unfolding to enrich the information in $V$. For brevity, we still utilize $V$ to represent the enriched features. 

The second variant is upsampling with only the pixel features, without utilizing the coordinates information, namely 
\begin{equation}
I_{p} = \Phi(\text{bilinear}(V_{p^*})),
\label{Eq:V-upsampler}
\end{equation}
where bilinear means the values $V\in \mathbb{R}^{H\times W\times D}$ are mapped to desired resolution by bilinear interpolation. After going through the pointwise MLP $\Psi$, the values $V$ are mapped to pixel values in HR space. As shown in Table \ref{table-upsampler}, the proposed upsampler greatly outperforms the two variants. This demonstrates that the proposed implicit transformer based feature modulation strategy is more effective than the direct mapping strategy.   

In addition, we also conduct ablation on the proposed periodic modulated function $\textit{sin}$. We replace it by \textit{tanh}, \textit{sigmoid}, and \textit{softmax} functions, respectively.
As shown in \tabref{reweigh_func}, 
utilizing \textit{sigmoid} or \textit{softmax} to perform nonlinear mapping, the performance drops more than (or around) 0.3 dB in both two test sets. The main reason is that \textit{sigmoid} and \textit{softmax} lead to positive values, which limit the ranges of modulated features. Correspondingly, \textit{tanh}, whose output range is $[-1, 1]$, generates better result than \textit{sigmoid} and \textit{softmax}. However, it is still lower than \textit{sin} function, which demonstrates that the periodic function contributes to high-frequency learning and performance improvement.








\subsubsection{Ablation on  Feature Extraction Backbone}
\begin{table}[h]
    \centering
    \caption{
		\textbf{Ablation on the proposed Dual-Branch Block}. The module \textit{"parallel"} is our proposed DBB, where conv-branch and attention-branch are in parallel. The variant \textit{"sequential"} means the two branches are stacked sequentially. \textit{"conv\ only"} means we only use the conv-branch in our DBB, and \textit{"attention\ only"} means we only adopt the attention-branch in DBB. All the results are evaluated on the SCID dataset for $\times4$ SR. 
}
    \vspace{-3mm}
    \label{tab:arch}
    \resizebox{0.48\textwidth}{!}
    {
    \begin{tabular}{l|cc|cc}
        \toprule[1.5pt]
        \multirow{2}{*}{\makecell[c]{Structure}} & 
    
        \multirow{2}{*}{Parameters}  & 
        \multirow{2}{*}{FLOPs}  & 
        \multirow{2}{*}{PSNR$\uparrow$} &  
        \multirow{2}{*}{SSIM$\uparrow$} 
         \\
         \\
         \hline
    Sequential   &  13.98 M & 320.87 G & 26.92 & 0.9033 \\
    Attention only  & 13.18 M  & 316.32 G & 27.38  &  0.9099 \\
    DWConv only  &  9.02 M & 277.23 G & 26.24   &  0.8942 \\
    \hline
    Parallel DWConv on input &  13.98 M & 320.87 G & 27.63   &  0.9133 \\
    Parallel DWConv on $V$ (Ours)  & 13.98 M & 320.87 G & 27.84 & 0.9149 \\
    \bottomrule[1.5pt]
    \end{tabular}
    }
\end{table}

Since our dual-branch block contains parallel attention and convolution branches, we replace it by transformer-block (attention branch + FFN), conv-block (conv-branch + FFN) and the sequential block, which is constructed by attention-branch+ conv-branch + FFN. For the three variants, the block number settings ($N$) and the ratio $\alpha$ are the same as that of the original setting for DBB. \tabref{arch} presents the comparison results. It can be observed that the proposed parallel solution (i.e., DBB) outperforms the three variants. Compared with the sequential block, the proposed DBB achieves 0.92 dB gain. \textcolor{red}{Compared with the conv block, which has larger parameters than the proposed DBB, we achieve nearly 1 dB gain.}

\begin{table*}[ht]
	\centering
	\caption{
		\textbf{  
  Quantitative comparison with state-of-the-art fixed-scale and continuous SR methods on SCI2K test set}. In-training-scale means the upsampling ratios are included in the training pairs and out-of-training-scale means the upsampling ratios are not ''seen" during training. The fixed-scale SR methods train different models for different upsampling ratios and the continuous SR methods train one model for all the upsampling ratios. All the models are trained on the training set of SCI2K. Values in \red{\textbf{red}} and \blue{\underline{blue}} indicate the best and the second best performance, respectively.
	}
	\vspace{-3mm}
	\label{tab:quality_val} 
	\resizebox{\textwidth}{!}{
    \begin{tabular}{l|ccc|ccccc}
    \toprule[1.5pt]
    \multirow{2}{*}{Methods} & \multicolumn{3}{c|}{In-training-scale} & \multicolumn{5}{c}{Out-of-training-scale} \\
     & $\times2$ & $\times3$ & $\times4$ & $\times6$ & $\times12$ & $\times18$ & $\times24$ & $\times30$\\
    \hline
    Bicubic & 28.32 / 0.9380 & 26.02 / 0.8912 & 24.81 / 0.8556 & 23.50 / 0.8116  & 21.67 / 0.7624 & 20.82 / 0.7514 & 20.22 / 0.7475 & 19.75 / 0.7450 \\
    EDSR \cite{lim2017edsr} & 38.68 / 0.9874 & 33.26 / 0.9713 & 29.84 / 0.9493 & -  & - & - & - & - \\
    RDN \cite{zhang2018RDN} & 38.68 / 0.9875 & 33.51 / 0.9716 & 30.25 / 0.9508  & - & - & - & - & - \\
    RCAN \cite{zhang2018rcan} & 40.18 / 0.9886 & 34.19 / 0.9734  & 30.75 / 0.9545 & - & - & - & - & -\\
    SwinIR\cite{liang21swinir} & \second{40.67} / \second{0.9890}   &  \second{34.79} / \second{0.9752} & \second{31.19} /  \second{0.9576} & - & - & - & - & -\\
    \midrule
    MetaSR \cite{hu2019metasr,chen2021liif} & 38.82 / 0.9875 & 33.24 / 0.9705 & 29.72 / 0.9467 & 26.36 / 0.8950  & 23.40 / 0.8187 & 22.28 / 0.7868 & 21.61 / 0.7708 & 21.08 / 0.7610 \\
    LIIF \cite{chen2021liif} & 38.85 / 0.9879  & 33.36 / 0.9719  & 29.98 / 0.9498 & 26.42 / 0.8986 & 23.44 / 0.8239 & 22.29 / 0.7913 & 21.61 / 0.7754 & 21.10 / 0.7654   \\
    ITSRN\cite{yang2021itsrn}  & 39.56 / 0.9883 & 34.11 / 0.9731 & 30.44 / 0.9520 & \second{26.61} / \second{0.9011} & \second{23.53} / {0.8243} & \second{22.36} / 0.7914  & \second{21.67} / 0.7746 & {21.15} / 0.7646 \\
    LTE\cite{lee2021lte} & 38.89 / 0.9879 & 33.28 / 0.9718 & 29.91 / 0.9497
    & 26.40 / 0.8998 & 23.48 / \second{0.8254} & 22.34 / \second{0.7929}  & 21.65 / \second{0.7760}  & \second{21.16} / \second{0.7663} \\
    \midrule
    \textbf{ITSRN++} & \first{41.26} / \first{0.9895} & \first{35.53} / \first{0.9768}  & \first{31.66} /
    \first{0.9597} & 
    \first{27.17} / \first{0.9118} & \first{23.86} / \first{0.8335}  & \first{22.58} / \first{0.7980} 
    & \first{21.86} / \first{0.7797}  &\first{21.33} / \first{0.7684}
    \\
    \bottomrule[1.5pt]
\end{tabular}
	}
	\vspace{-3mm}
\end{table*}

\begin{table*}[ht]
	\centering
	\caption{
		\textbf{Quantitative comparison with state-of-the-art fixed-scale and continuous SR methods on three screen content quality assessment datasets. }. All the models are trained on the training set of SCI2K. Values in \red{\textbf{red}} and \blue{\underline{blue}} indicate the best and the second best performance, respectively.}. 

	\vspace{-3mm}
	\label{tab:quality_benchmark} 
	\resizebox{\textwidth}{!}{
    \begin{tabular}{c|c|ccc|ccc}
        \toprule[1.5pt]
        \multirow{2}{*}{Dataset} & \multirow{2}{*}{Method} & \multicolumn{3}{c|}{In-training-scale} & \multicolumn{3}{c}{Out-of-training-scale} \\
        & & $\times$2 & $\times$3 & $\times$4  & $\times$6  & $\times$8 &  $\times$10\\
        \midrule
        \multirow{5}{*}[-16pt]{CCT\cite{min2017CCT}} 
        & EDSR~\cite{lim2017edsr}  & 31.62 / 0.9630 & 27.14 / 0.9171  & 24.67 / 0.8619  & -- & -- & -- \\
        & RDN~\cite{zhang2018RDN} & 31.68 / 0.9648 & 27.26 / 0.9158 & 24.73 / 0.8626 & -- & -- & -- \\
        & RCAN~\cite{zhang2018rcan} & 32.79 / 0.9703 & {27.74} / 0.9234 & 25.22 / 0.8770 & -- & -- & -- \\
        & SwinIR~\cite{liang21swinir} & \second{33.20} / \second{0.9710} & \second{28.28} / \second{0.9332} & \second{25.50} / \second{0.8862} & -- & -- & -- \\
        & MetaSR~\cite{hu2019metasr} & 31.91 / 0.9665 & 27.38 / 0.9195 & 24.57 / 0.8597  & 22.24 / 0.7641  & 21.24 / 0.6990  & 20.71 / 0.6669\\
        & LIIF~\cite{chen2021liif}  & 32.26 / 0.9684 & 27.63 / 0.9240  & 24.75 / 0.8634 & 
        22.31 / 0.7699 & 
        21.27 / 0.7050 & 
        20.72 / 0.6723 \\
        & ITSRN\cite{yang2021itsrn}  & 32.56 / 0.9697 & {28.00} / {0.9267} & 25.07 / 0.8705 
        & \second{22.38} / 0.7719 
        &  21.35 / 0.7062 
        & 20.79 / 0.6727 \\
        & LTE~\cite{lee2021lte}  & 32.51 / 0.9709 & 27.73 / 0.9282 & 24.87 / 0.8731 
        & \second{22.38} / \second{0.7781} 
        & \second{21.37} / \second{0.7139} 
        & \second{20.84} / \second{0.6781} \\
        & \textbf{ITSRN++}  & \first{34.23} / \first{0.9781} & \first{29.30} / \first{0.9452} & \first{26.29 }/ \first{0.9017} 
        & \first{23.06} / \first{0.8067}  
        & \first{21.65} / \first{0.7284} 
        & \first{20.87} / \first{0.6795} 
        \\
        \midrule
        \multirow{5}{*}[-16pt]{SCID~\cite{SCID2017}} 
        & EDSR~\cite{lim2017edsr} & 33.92 / 0.9700 & 29.00 / 0.9323 & 25.85 / 0.8873 & -- & -- & -- \\
        & RDN~\cite{zhang2018RDN} & 34.00 / 0.9713 & 29.13 / 0.9317 & 26.11 / 0.8906 & -- & -- & -- \\
        & RCAN~\cite{zhang2018rcan} & 34.90 / 0.9743 & 29.65 / 0.9358 & 26.59 / 0.8989 & -- & -- & --\\
        & SwinIR~\cite{liang21swinir} & \second{35.54} / \second{0.9751} & \second{30.29} / \second{0.9419} & \second{26.89} / \second{0.9037} & -- & -- & --\\
        & MetaSR~\cite{hu2019metasr} & 34.11 / 0.9719 & 29.24 / 0.9331 & 25.99 / 0.8891 
        & 22.65 / 0.7941 
        &  21.25 / 0.7238  
        & 20.58 / 0.6867 \\
        & LIIF~\cite{chen2021liif}  & 34.22 / 0.9728 & 29.38 / 0.9350 & 26.22 / 0.8932 
        & 22.69 / 0.8007 
        & 21.29 / 0.7315 
        & 20.59 / 0.6925 \\
        & ITSRN~\cite{yang2021itsrn}  & 34.67 / 0.9739 & {29.98} / {0.9378} & 26.68 / 0.8974 
        & \second{22.86} / 0.8029 
        & 21.36 / 0.7335  
        & 20.65 / 0.6933\\
        & LTE~\cite{lee2021lte}  & 34.54 / 0.9746 & 29.54 / 0.9383 & 26.24 / 0.8972  
        & 22.74 / \second{0.8053}  
        & \second{21.38} / \second{0.7389} 
        & \second{20.67} / \second{0.6980}\\
        & \textbf{ITSRN++}  & \first{36.07} / \first{0.9788} & \first{31.14} / \first{0.9487}  & \first{27.84} / \first{0.9149} 
        & \first{23.63} / \first{0.8324} 
        & \first{21.65} / \first{0.7478} 
        & \first{20.71} / \first{0.6948}\\
        \midrule

        \multirow{5}{*}[-16pt]{SIQAD~\cite{yang2015SIQAD}} 
        & EDSR~\cite{lim2017edsr} & 33.58 / 0.9806 & 27.95 / 0.9462 & 23.95 / 0.8891 & -- & -- & -- \\
        & RDN~\cite{zhang2018RDN} & 33.68 / 0.9807 & 28.07 / 0.9455 & 24.37 / 0.8930 & -- & -- & -- \\
        & RCAN~\cite{zhang2018rcan} & 34.79 / 0.9828 & 29.02 / 0.9515 & 25.01 / 0.9040 & -- & -- & --\\
        & SwinIR~\cite{liang21swinir} & \second{35.35} / 0.9829 & \second{29.57} / \second{0.9553} & \second{25.39} / \second{0.9102} & -- & -- & --\\
        & MetaSR~\cite{hu2019metasr} & 34.13 / 0.9818  & 28.42 / 0.9479  & 23.82 / 0.8831 
        & 20.24 / 0.7418 
        & 19.34 / 0.6729 
        & 18.77 / 0.6263\\
        & LIIF~\cite{chen2021liif}  & 34.23 / 0.9821 & 28.55 / 0.9494 & 24.09 / 0.8899 
        & 20.26 / 0.7479 
        & 19.34 / 0.6772 
        & 18.73 / 0.6309 \\
        & ITSRN~\cite{yang2021itsrn}  & {34.92} / \second{0.9832} & 29.56 / {0.9541} & 24.70 / 0.8981 
        & \second{20.38} / 0.7536 
        & \second{19.41} / 0.6811 
        & \second{18.82} / 0.6340 \\
        & LTE~\cite{lee2021lte}  & 34.52 / 0.9830 & 28.54 / 0.9505  & 24.01 / 0.8919 
        & 20.29 / \second{0.7538} 
        & 19.39 / \second{0.6816} 
        & \second{18.82} / \second{0.6354}\\
        & \textbf{ITSRN++}  & \first{36.06} / \first{0.9852}  & \first{31.11} / \first{0.9618}  & \first{26.47} / \first{0.9220} 
        & \first{20.98} / \first{0.7869}  
        & \first{19.62} / \first{0.6947}  
        & \first{18.88} / \first{0.6346}\\
        \bottomrule[1.5pt]
    \end{tabular} 
	}
	\vspace{-3mm}
\end{table*}
\begin{table}
	\centering
	\caption{
		Quantitative evaluation on SCI2K-compression dataset for $\times4$ SR. All the models are trained on the training set of SCI2K-compression and evaluated on the compressed test set of SCI2K, with the compression quality factor setting to 75, 85, and 95 respectively. Values in \red{\textbf{red}} and \blue{\underline{blue}} indicate the best and the second best performance, respectively.
	}
	\vspace{-3mm}
	\label{tab:quality_benchmark_compress} 
	\resizebox{0.48\textwidth}{!}{
    \begin{tabular}{c|ccc}
        \toprule
        \multirow{2}{*}{Method} & \multicolumn{3}{c}{Compression Factor}  \\
        & $q=75$ & $q=85$ & $q=95$ \\
        \hline
        RCAN~\cite{zhang2018rcan} & 27.11 / 0.9093 & 27.83 / 0.9210 & 28.61 / 0.9335 \\
        SwinIR~\cite{liang21swinir} & \second{27.34} / \second{0.9134} & \second{28.10} / \second{0.9250} & \second{28.93} / \second{0.9375} \\
        MetaSR~\cite{hu2019metasr} & 26.93 / 0.9029 & 27.41 / 0.9142 & 28.02 / 0.9262 \\
        LIIF~\cite{chen2021liif}  & 26.90 / 0.9035 & 27.53 / 0.9154  & 28.18 / 0.9271  \\
        ITSRN\cite{yang2021itsrn}  & 27.11 / 0.9076 & 27.82 / 0.9197 & 28.56 / 0.9318  \\
        LTE~\cite{lee2021lte}  & 26.94 / 0.9057 & 27.59 / 0.9175 & 28.32 / 0.9302 \\
        \textbf{ITSRN++}  & \first{27.60} / \first{0.9168} & \first{28.40} / \first{0.9289} & \first{29.34} / \first{0.9419} 
        \\
        \bottomrule
    \end{tabular} 
	}
	\vspace{-3mm}
 \label{table:comp}
\end{table}

\figref{ablation_dpfb} presents the visual comparison results and the corresponding local attribution maps (LAM) \cite{gu202LAM} of the three variants and our method. The first row presents the LAM results, where the red points denote pixels that contribute to the reconstruction. It can be observed that when only using convolution, the region of LAM is the smallest. It indicates that the convolution operation is good at utilizing local correlations due to its inductive bias. The attention-only variant has a larger LAM since self-attention can explore long-range correlations, and it achieves better results than the conv-only variant. The sequential connection variant has the largest LAM region. However, many uncorrelated pixels are also involved and this leads to worse results compared with the attention-only variant. In contrast, our proposed parallel solution leads to a larger LAM region, and the pixels inside LAM are all correlated with the target patch. Therefore, the proposed parallel strategy achieves the best reconstruction result.  




\begin{figure*}
	\setlength{\fs}{-0.4cm}
	\scriptsize
	\centering
	\begin{tabular}{cc}
	\hspace{-0.4cm}
	\begin{adjustbox}{valign=t}
	\begin{tabular}{c}
	\includegraphics[width=0.18\textwidth]{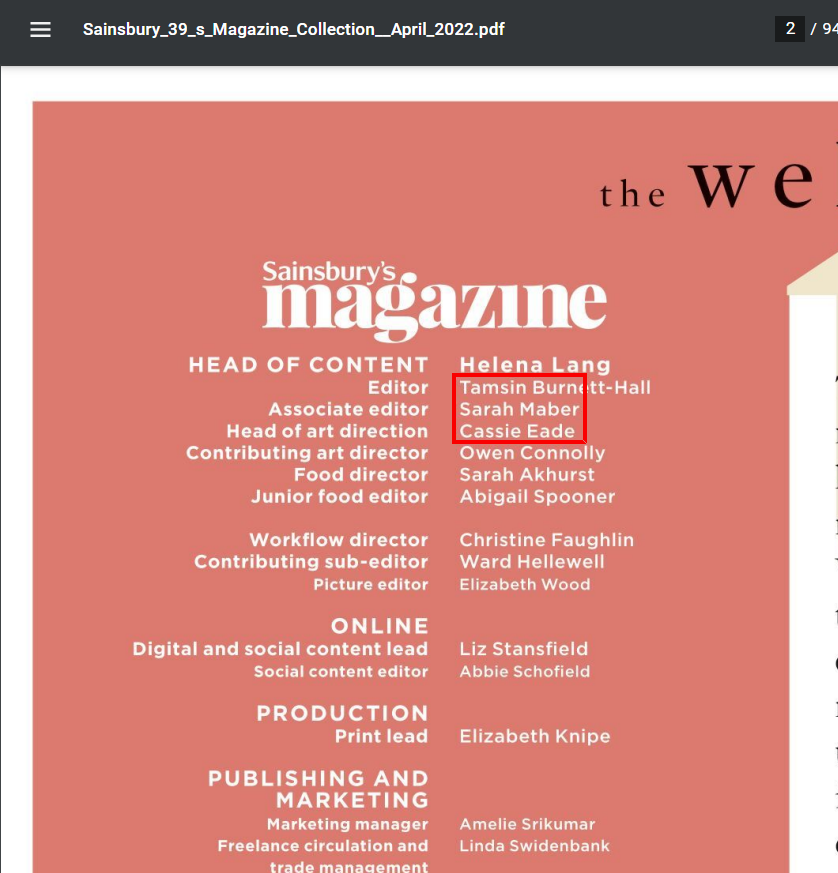}
	\\
	Ground-truth HR
	\\
	\textsc{SCI2K}: img1999
	\end{tabular}
	\end{adjustbox}
	\hspace{-0.46cm}
	\begin{adjustbox}{valign=t}
	\begin{tabular}{ccccc}
	\includegraphics[width=0.15\textwidth]{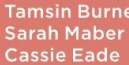} \hspace{\fs} &
	\includegraphics[width=0.15\textwidth]{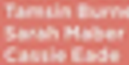} \hspace{\fs} &
	\includegraphics[width=0.15\textwidth]{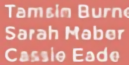} \hspace{\fs} &
	\includegraphics[width=0.15\textwidth]{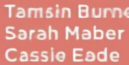} \hspace{\fs} &
	\includegraphics[width=0.15\textwidth]{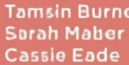} \hspace{\fs} 
    \vspace{1mm}
	\\
	HR \hspace{\fs} &
	Bicubic \hspace{\fs} &
	RDN~\cite{zhang2018RDN} \hspace{\fs} &
	RCAN~\cite{zhang2018rcan} \hspace{\fs} &
	SwinIR~\cite{liang21swinir} \hspace{\fs} 

	\vspace{1mm}
	\\
	\includegraphics[width=0.15\textwidth]{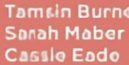} \hspace{\fs} &
	\includegraphics[width=0.15\textwidth]{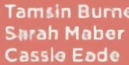} \hspace{\fs} &
	\includegraphics[width=0.15\textwidth]{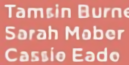} \hspace{\fs} &
	\includegraphics[width=0.15\textwidth]{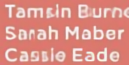} \hspace{\fs} &
	\includegraphics[width=0.15\textwidth]{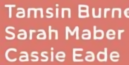}  \hspace{\fs} 
    \vspace{1mm}
	\\ 
	MetaSR~\cite{hu2019metasr} \hspace{\fs} &
	LIIF~\cite{chen2021liif} \hspace{\fs} &
	LTE~\cite{lee2021lte} \hspace{\fs} &
	ITSRN~\cite{yang2021itsrn} \hspace{\fs} &
	ITSRN++(ours) \hspace{\fs} 


	\\
	\end{tabular}
	\end{adjustbox}
	\vspace{2mm}
	\\
	\hspace{-0.4cm}
	\begin{adjustbox}{valign=t}
	\begin{tabular}{c}
	\includegraphics[width=0.18\textwidth]{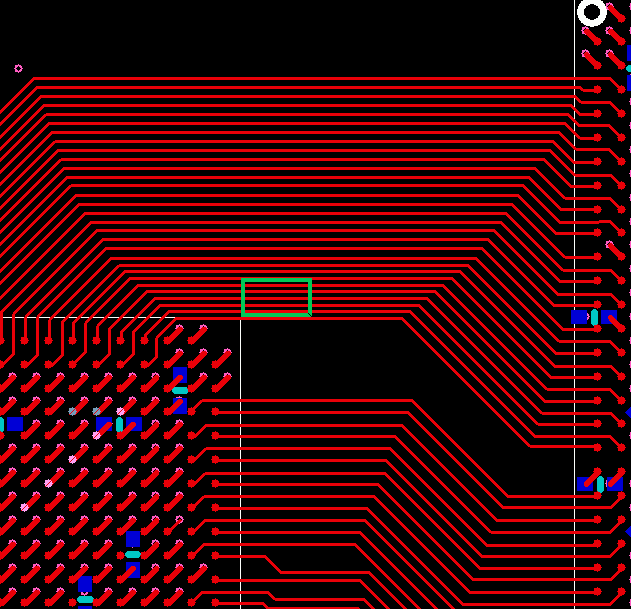}
	\\
	Ground-truth HR
	\\
	\textsc{CCT}: img023
	\end{tabular}
	\end{adjustbox}
	\hspace{-0.46cm}
	\begin{adjustbox}{valign=t}
	\begin{tabular}{ccccc}
	\includegraphics[width=0.15\textwidth]{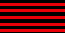} \hspace{\fs} &
	\includegraphics[width=0.15\textwidth]{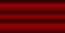} \hspace{\fs} &
	\includegraphics[width=0.15\textwidth]{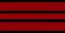} \hspace{\fs} &
	\includegraphics[width=0.15\textwidth]{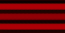} \hspace{\fs} &
	\includegraphics[width=0.15\textwidth]{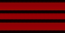} \hspace{\fs} 
    \vspace{0.5mm}
	\\
	HR \hspace{\fs} &
	Bicubic \hspace{\fs} &
	RDN~\cite{zhang2018RDN} \hspace{\fs} &
	RCAN~\cite{zhang2018rcan} \hspace{\fs} &
	SwinIR~\cite{liang21swinir} \hspace{\fs} 
    \vspace{0.5mm}
	
 
	\\
	\includegraphics[width=0.15\textwidth]{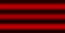} \hspace{\fs} &
	\includegraphics[width=0.15\textwidth]{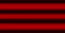} \hspace{\fs} &
	\includegraphics[width=0.15\textwidth]{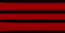} \hspace{\fs} &
	\includegraphics[width=0.15\textwidth]{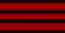} \hspace{\fs} &
	\includegraphics[width=0.15\textwidth]{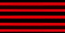} \hspace{\fs} 
	\vspace{0.5mm}
	\\ 
	MetaSR~\cite{hu2019metasr} \hspace{\fs} &
	LIIF~\cite{chen2021liif} \hspace{\fs} &
	LTE~\cite{lee2021lte} \hspace{\fs} &
	ITSRN~\cite{yang2021itsrn} \hspace{\fs} &
	ITSRN++(Ours)\hspace{\fs} 
	\\
	\end{tabular}
	\end{adjustbox}
	\vspace{2mm}
	\\
	\hspace{-0.4cm}
	\begin{adjustbox}{valign=t}
	\begin{tabular}{c}
	\includegraphics[width=0.18\textwidth]{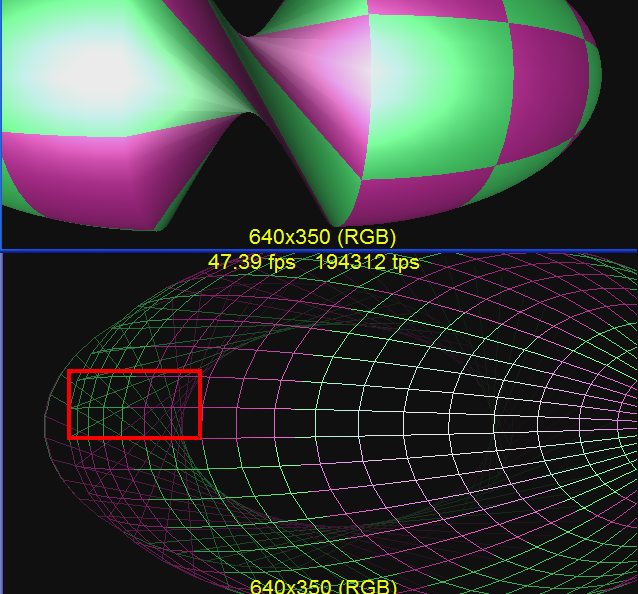}
	\\
	Ground-truth HR
	\\
	\textsc{SCID}: img032
	\end{tabular}
	\end{adjustbox}
	\hspace{-0.46cm}
	\begin{adjustbox}{valign=t}
	\begin{tabular}{ccccc}
	\includegraphics[width=0.15\textwidth]{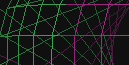} \hspace{\fs} &
	\includegraphics[width=0.15\textwidth]{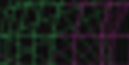} \hspace{\fs} &
	\includegraphics[width=0.15\textwidth]{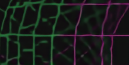} \hspace{\fs} &
	\includegraphics[width=0.15\textwidth]{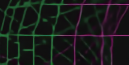} \hspace{\fs} &
	\includegraphics[width=0.15\textwidth]{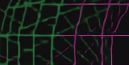} \hspace{\fs} 
    \vspace{0.5mm}
	\\
	HR \hspace{\fs} &
	Bicubic \hspace{\fs} &
	RDN~\cite{zhang2018RDN} \hspace{\fs} &
	RCAN~\cite{zhang2018rcan} \hspace{\fs} &
	SwinIR~\cite{liang21swinir} \hspace{\fs} 
    \vspace{0.5mm}
	
	\\
	\includegraphics[width=0.15\textwidth]{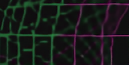} \hspace{\fs} &
	\includegraphics[width=0.15\textwidth]{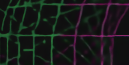} \hspace{\fs} &
	\includegraphics[width=0.15\textwidth]{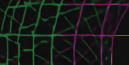} \hspace{\fs} &
	\includegraphics[width=0.15\textwidth]{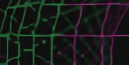} \hspace{\fs} &
	\includegraphics[width=0.15\textwidth]{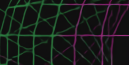} \hspace{\fs} 
	\vspace{0.5mm}
	\\ 
	MetaSR~\cite{hu2019metasr} \hspace{\fs} &
	LIIF~\cite{chen2021liif} \hspace{\fs} &
	LTE~\cite{lee2021lte} \hspace{\fs} &
	ITSRN~\cite{yang2021itsrn} \hspace{\fs} &
	ITSRN++(Ours)\hspace{\fs} 
	\\
	\end{tabular}
	\end{adjustbox}
	
	\vspace{2mm}
	\\
	\hspace{-0.4cm}
	\begin{adjustbox}{valign=t}
	\begin{tabular}{c}
	\includegraphics[width=0.18\textwidth]{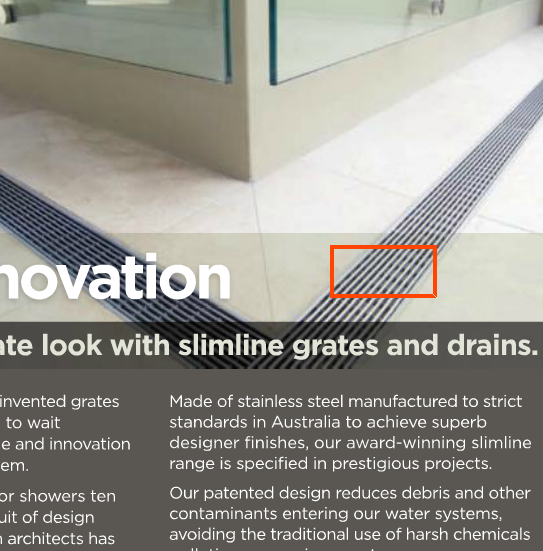}
	\\
	Ground-truth HR
	\\
	\textsc{SIQAD}: img017
	\end{tabular}
	\end{adjustbox}
	\hspace{-0.46cm}
	\begin{adjustbox}{valign=t}
	\begin{tabular}{ccccc}
	\includegraphics[width=0.15\textwidth]{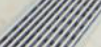} \hspace{\fs} &
	\includegraphics[width=0.15\textwidth]{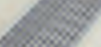} \hspace{\fs} &
	\includegraphics[width=0.15\textwidth]{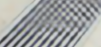} \hspace{\fs} &
	\includegraphics[width=0.15\textwidth]{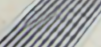} \hspace{\fs} &
	\includegraphics[width=0.15\textwidth]{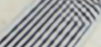} \hspace{\fs} 
    \vspace{0.5mm}
	\\
	HR \hspace{\fs} &
	Bicubic \hspace{\fs} &
	RDN~\cite{zhang2018RDN} \hspace{\fs} &
	RCAN~\cite{zhang2018rcan} \hspace{\fs} &
	SwinIR~\cite{liang21swinir} \hspace{\fs} 
    \vspace{0.5mm}
	

	\\
	\includegraphics[width=0.15\textwidth]{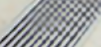} \hspace{\fs} &
	\includegraphics[width=0.15\textwidth]{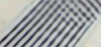} \hspace{\fs} &
	\includegraphics[width=0.15\textwidth]{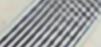} \hspace{\fs} &
	\includegraphics[width=0.15\textwidth]{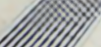} \hspace{\fs} &
	\includegraphics[width=0.15\textwidth]{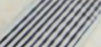} \hspace{\fs} 
	\vspace{0.5mm}
	\\ 
	MetaSR~\cite{hu2019metasr} \hspace{\fs} &
	LIIF~\cite{chen2021liif} \hspace{\fs} &
	LTE~\cite{lee2021lte} \hspace{\fs} &
	ITSRN~\cite{yang2021itsrn} \hspace{\fs} &
	ITSRN++(Ours)\hspace{\fs} 

	\\
	\end{tabular}
	\end{adjustbox}
	\end{tabular}
	\caption{
		\textbf{Visual comparison for $\times4$ SR on the \textsc{SCI2K}, \textsc{CCT}, \textsc{SIQAD}, and \textsc{SCID} datasets.}
	}
	\label{fig:result_4x_sci}
\end{figure*}

\begin{table*}
	\centering
	\caption{
		\textbf{Quantitative evaluation of state-of-the-art SR methods on five benchmark SR datasets. All the models are trained on the DIV2K dataset.}. The dashed line is used to separate fixed-scale and continuous SR methods.  
The results for fixed-scale SR methods are generated with models trained with $\times2, \times3, \times4$ and $\times8$ LR-HR pairs, respectively. The results for continuous SR methods are generated with one model trained with continuous random scales uniformly sampled from $\times1$~$\times4$ LR-HR pairs. All results are quoted from their original papers, except MetaSR and LIIF (indicated by $^\dag$). For them, we utilize the results reported in \cite{lee2021lte}, which changes their feature extraction backbone to SwinIR.    Since CSNLN~\cite{Mei2020CSNLN}, IGNN\cite{zhou2020IGNN} NSLN~\cite{Mei_2021_NLSN}, and ENLCN~\cite{xia2022ENLCA}  did not  report their $\times 8$ results, we omit them. Values in \red{\textbf{red}} and \blue{\underline{blue}} indicate the best and the second best performance, respectively. 
	}
	\vspace{-3mm}
	\label{tab:quality_benchmark_natual} 
	\resizebox{\textwidth}{!}{
	\begin{tabular}{lccccccc}
        \hline
	\multirow{2}{*}{Method} & \multirow{2}{*}{Year} & \multirow{2}{*}{Scale} &
		\textsc{Set5} & \textsc{Set14} & \textsc{BSDS100} & \textsc{Urban100} & \textsc{Manga109} \\
  
		&
            &
            &
		PSNR / SSIM  & 
		PSNR / SSIM  & 
		PSNR / SSIM  & 
		PSNR / SSIM  & 
		PSNR / SSIM \\
        \hline
        \hline
		Bicubic & -- &$\times$2  &
                33.66 / 0.9299 &
                30.24 / 0.8688 &
                29.56 / 0.8431 &
                26.88 / 0.8403 &
                30.80 / 0.9339
		\\
		EDSR~\cite{lim2017edsr} & 2017 & $\times$2 &
                38.11 / 0.9602 &
                33.92 / 0.9195 &
                32.32 / 0.9013 &
                32.93 / 0.9351 &
                39.10 / 0.9773
		\\
		RCAN~\cite{zhang2018rcan} & 2018 & $\times$2  &
                38.27 / 0.9614  &
                34.12 / 0.9216  &
                32.41 / 0.9027  &
                33.34 / 0.9384  &
                39.44 / 0.9786
		\\
            CSNLN~\cite{Mei2020CSNLN}  & 2020 &$\times$2  &
                38.28 / 0.9616&
                34.12 / 0.9223 &
                32.40 / 0.9024 &
                33.25 / 0.9386 &
                39.37 / 0.9785
		\\
            IGNN~\cite{zhou2020IGNN}  & 2020 &$\times$2  &
                38.24 / 0.9613 &
                34.07 / 0.9217 &
                32.41 / 0.9025 &
                33.23 / 0.9383 &
                39.35 / 0.9786
            \\
            HAN~\cite{niu2020HAN}& 2020 & $\times$2  &
                38.27 / 0.9614 &
                34.16 / 0.9217 &
                32.41 / 0.9027 &
                33.35 / 0.9385 &
                39.46 / 0.9785
            \\
            NLSN~\cite{Mei_2021_NLSN} & 2021 & $\times$2  &
                38.34 / 0.9618 &
                34.08 / {0.9231} &
                32.43 / 0.9027 &
                33.42 / {0.9394} &
                39.59 / 0.9789
            \\
		SwinIR~\cite{liang21swinir} & 2021 & $\times$2  &
    		\second{38.35} / \first{0.9620} &
    		{34.14} / 0.9227  &
    		32.44 / 0.9030 &
    		33.40 / 0.9393 &
    		{39.60} / \first{0.9792}
            \\	
       
            ENLCN~\cite{xia2022ENLCA}& 2022 & $\times$2  &
                \first{38.37} / 0.9618 &
                {34.17} / 0.9229 &
                \first{32.49} / \first{0.9032} &
                \second{33.56} / \second{0.9398} &
                \second{39.64} / \second{0.9791}
            \\
        \hdashline
         SwinIR-MetaSR$^\dag$~\cite{liang21swinir,hu2019metasr} & 2019 & $\times$2  &
    		38.12 / 0.9614 &
    		34.09 / 0.9227  &
    		32.35 / 0.9021 &
    		33.23 / 0.9387 &
    		39.29 / 0.9784
            \\	
           SwinIR-LIIF$^\dag$~\cite{liang21swinir,chen2021liif} & 2021 & $\times$2  &
    		38.15 / 0.9615 &
    		34.09 / 0.9223 &
    		32.35 / 0.9020 &
    		33.31 / 0.9389 &
    		39.38 / 0.9786 
            \\
           SwinIR-LTE~\cite{lee2021lte} & 2022 & $\times$2  &
    		38.21 / 0.9616 &
    		\second{34.19} / \second{0.9234}  &
    		32.39 / 0.9026 &
    		33.45 / {0.9405} &
    		39.44 / 0.9786
            \\
	\textbf{ITSRN++} & -- & $\times$2  &
    		38.30 / {0.9619}  &
    		\first{34.41} / \first{0.9236} &
    		\second{32.45} / \first{0.9032}  &
    		\first{33.79} / \first{0.9423}   &
    		\first{39.69} / \second{0.9791}
		\\
        
	\hline
        \hline

	Bicubic & -- & $\times$3  &
            30.39 / 0.8682 &
            27.55 / 0.7742 &
            27.21 / 0.7385 &
            24.46 / 0.7349 &
            26.95 / 0.8556
		\\
	EDSR~\cite{lim2017edsr} & 2017 & $\times$3  &
            34.65 / 0.9280 &
            30.52 / 0.8462 &
            29.25 / 0.8093 &
            28.80 / 0.8653 &
            34.17 / 0.9476
		\\
        RCAN~\cite{zhang2018rcan} & 2018& $\times$3  &
            34.74 / 0.9299 &
            30.65 / 0.8482 &
            29.32 / 0.8111 &
            29.09 / 0.8702 &
            34.44 / 0.9499
		\\
	CSNLN~\cite{Mei2020CSNLN} & 2020& $\times$3  &
            34.74 / 0.9300 &
            30.66 / 0.8482 &
            29.33 / 0.8105 &
            29.13 / 0.8712 &
            34.45 / 0.9502
		\\
	IGNN~\cite{zhou2020IGNN} & 2020& $\times$3  &
            34.72 / 0.9298 &
            30.66 / 0.8484 &
            29.31 / 0.8105 &
            29.03 / 0.8696 &
            34.39 / 0.9496
		\\
	HAN~\cite{niu2020HAN} & 2020 &$\times$3  &
            34.75 / 0.9299 &
            30.67 / 0.8483 &
            29.32 / 0.8110 &
            29.10 / 0.8705 &
            34.48 / 0.9500
		\\
	NLSN~\cite{Mei_2021_NLSN} & 2021  &$\times$3 &
            34.85 / 0.9306 &
            30.70 / 0.8485 &
            29.34 / 0.8117 &
            29.25 / 0.8726 &
            34.57 / 0.9508
		\\
	SwinIR~\cite{liang21swinir}  & 2021&$\times$3  &
            \second{34.89} / \second{0.9312} &
            {30.77} / {0.8503} &
            {29.37} / {0.8124} &
            {29.29} / {0.8744} &
            {34.74} / {0.9518}
        \\
        ENLCN~\cite{xia2022ENLCA}& 2022 & $\times$3  &
		-- &
		-- &
		-- &
		-- &
		--
	\\
        \hdashline
        SwinIR-MetaSR$^\dag$~\cite{liang21swinir,hu2019metasr} & 2019 & $\times$3  &
		34.70 / 0.9300 &
		30.64 / 0.8498  &
		29.29 / 0.8115 &
		29.10 / 0.8728 &
		34.57 / 0.9508
       \\	
       SwinIR-LIIF$^\dag$~\cite{liang21swinir,chen2021liif} & 2021& $\times$3  &
		34.81 / 0.9306 &
		30.72 / 0.8503  &
		29.33 / 0.8123&
		29.31 / 0.8753 &
		34.64 / 0.9514   
        \\
       SwinIR-LTE~\cite{lee2021lte} & 2022 & $\times$3  &
		34.83 / 0.9309 &
		\second{30.77} / \second{0.8513}  &
		\second{29.37} / \second{0.8128} &
		\second{29.40} / \second{0.8768} &
		\second{34.75} / \second{0.9519}
        \\
        
	\textbf{ITSRN++} & --&$\times$3   &
            \first{34.90} / \first{0.9315} &
            \first{30.94} / \first{0.8523} &
            \first{29.40} / \first{0.8137} &
            \first{29.67} / \first{0.8802} &
            \first{35.02} / \first{0.9529}
	\\
	\hline
         \hline
	Bicubic & -- & $\times$4  &
            28.42 / 0.8104 &
            26.00 / 0.7027 &
            25.96 / 0.6675 &
            23.14 / 0.6577 &
            24.89 / 0.7866 
		\\
	EDSR~\cite{lim2017edsr} & 2017 & $\times$4  &
            32.46 / 0.8968 &
            28.80 / 0.7876 &
            27.71 / 0.7420 &
            26.64 / 0.8033 &
            31.02 / 0.9148
		\\
	RCAN~\cite{zhang2018rcan} & 2018 & $\times$4  &
            32.63 / 0.9002 & 
            28.87 / 0.7889 &
            27.77 / 0.7436 &
            26.82 / 0.8087 &
            31.22 / 0.9173
		\\
	CSNLN~\cite{Mei2020CSNLN} & 2020 & $\times$4  &
            32.68 / 0.9003 &
            28.95 / 0.7888 &
            27.80 / 0.7439 &
            {27.22} / {0.8168} &
            31.43 / 0.9201
		\\
	IGNN~\cite{zhou2020IGNN} & 2020 & $\times$4  &
            32.57 / 0.8998 &
            28.85 / 0.7891 &
            27.77 / 0.7434 &
            26.84 / 0.8090 &
            31.28 / 0.9182
		\\
	HAN~\cite{niu2020HAN} & 2020 & $\times$4  &
            32.64 / 0.9002 &
            28.90 / 0.7890 &
            27.80 / 0.7442 &
            26.85 / 0.8094 &
            31.42 / 0.9177
		\\
	NLSN~\cite{Mei_2021_NLSN} & 2021 & $\times$4  &
            32.59 / 0.9000 &
            28.87 / 0.7891 &
            27.78 / 0.7444 &
            26.96 / 0.8109 &
            31.27 / 0.9184
		\\
	SwinIR~\cite{liang21swinir} & 2021 &$\times$4  &
            \second{32.72} / {0.9021} &
            {28.94} / {0.7914} &
            {27.83} / {0.7459} &
            27.07 / 0.8164 &
            {31.67} / {0.9226}
        \\
        ENLCN~\cite{xia2022ENLCA}& 2022 & $\times$4  &
            32.67 / 0.9004 &
            {28.94} / 0.7892 &
            {27.82} / {0.7452} &
            {27.12} / {0.8141} &
            {31.33} / {0.9188}
        \\
        \hdashline
        SwinIR-MetaSR$^\dag$~\cite{liang21swinir,hu2019metasr} & 2019 & $\times$4  & 
		32.43 / 0.8985 &
		28.83 / 0.7898  &
		27.74 / 0.7440 &
		26.75 / 0.8094 &
		31.34 / 0.9187
       \\	
       SwinIR-LIIF$^\dag$~\cite{liang21swinir,chen2021liif} 
       & 2021 & $\times$4  &
		32.69 / 0.9017 &
		28.96 / 0.7914  &
		27.82 / 0.7457   &
		27.13 / 0.8164 &
		31.69 / 0.9222
        \\
       SwinIR-LTE~\cite{lee2021lte} & 2022 & $\times$4  &
		\first{32.77} / \first{0.9024} &
		\second{29.04} / \second{0.7927}  &
		\second{27.85} / \second{0.7467} &
		\second{27.23} / \second{0.8195} &
		\second{31.78} / \second{0.9234}
        \\
	\textbf{ITSRN++} & -- & $\times$4  &
            \second{32.72} / \second{0.9023} &
            \first{29.07} / \first{0.7932} &
            \first{27.87} / \first{0.7473}&
            \first{27.45} / \first{0.8234}&
            \first{32.01} / \first{0.9249}
	\\
	\hline
        \hline
	Bicubic & -- & $\times$8   &
            24.40 / 0.6580 &
            23.10 / 0.5660 &
            23.67 / 0.5480 &
            20.74 / 0.5160 &
            21.47 / 0.6500 
	\\
	EDSR~\cite{lim2017edsr} & 2017 &$\times$8  &
            26.96 / 0.7762 &
            24.91 / 0.6420 &
            24.81 / 0.5985 &
            22.51 / 0.6221 &
            24.69 / 0.7841
	\\
	RCAN~\cite{zhang2018rcan} & 2018 & $\times$8  &
        27.31 / 0.7878 &
        25.23 / 0.6511 &
        24.98 / 0.6058 &
        23.00 / 0.6452 &
        25.24 / 0.8029
	\\
	CSNLN~\cite{Mei2020CSNLN} & 2020 &$\times$8   &
        -- &
		-- &
		-- &
		-- &
		--
		\\
	IGNN~\cite{zhou2020IGNN} & 2020 &$\times$8  &
		-- &
		-- &
		-- &
		-- &
		--
		\\
	HAN~\cite{niu2020HAN} & 2020 & $\times$8 &
		27.33 / \first{0.7884} &
		25.24 / 0.6510 &
		24.98 / 0.6059 &
		22.98 / 0.6347 &
		25.20 / 0.8011
		\\
	NLSN~\cite{Mei_2021_NLSN} & 2021&$\times$8  &
		-- &
		-- &
		-- &
		-- &
		--
		\\
	SwinIR~\cite{liang21swinir} & 2021 &$\times$8  &
            {27.37} / {0.7877}&
            {25.26} / {0.6523}&
            {24.99} / {0.6063}&
            23.03 / 0.6457&
            25.26 / 0.8005
		\\
        ENLCN~\cite{xia2022ENLCA}& 2022& $\times$8  &
		-- &
		-- &
		-- &
		-- &
		--
	    \\
        \hdashline
        SwinIR-MetaSR$^\dag$~\cite{liang21swinir,hu2019metasr} & 2019 & $\times$8 &
		27.18 / 0.7749 &
		25.20 / 0.6456 &
		  24.93 / 0.6019 &
		22.92 / 0.6361 &
		24.95 / 0.7838
       \\	
       SwinIR-LIIF$^\dag$~\cite{liang21swinir,chen2021liif} 
       & 2021& $\times$8  &
		27.32 / 0.7836 &
		25.31 / 0.6505  &
		24.99 / 0.6052 &
		23.11 / 0.6477   &
		25.27 / 0.7996
        \\
       SwinIR-LTE~\cite{lee2021lte} & 2022 & $\times$8  &
		\second{27.36} / 0.7864 &
		\first{25.39} / \first{0.6529}  &
	    \second{25.02} / \second{0.6067}  &
		  \second{23.16} / \second{0.6510} &
		\second{25.42} / \second{0.8043} 
        \\
	\textbf{ITSRN++} & -- & $\times$8 &
            \first{27.39} / \second{0.7865} &
            \second{25.36} / \second{0.6524} &
            \first{25.03} / \first{0.6070} &
            \first{23.28} / \first{0.6561} &
            \first{25.58} / \first{0.8080}
	\\
	\hline
	\end{tabular}
	}
	\vspace{-3mm}
\end{table*}
\begin{figure*}
	\setlength{\fs}{-0.4cm}
	\scriptsize
	\centering
	\begin{tabular}{cc}
	\hspace{-0.4cm}
	\begin{adjustbox}{valign=t}
	\begin{tabular}{c}
	\includegraphics[width=0.2\textwidth]{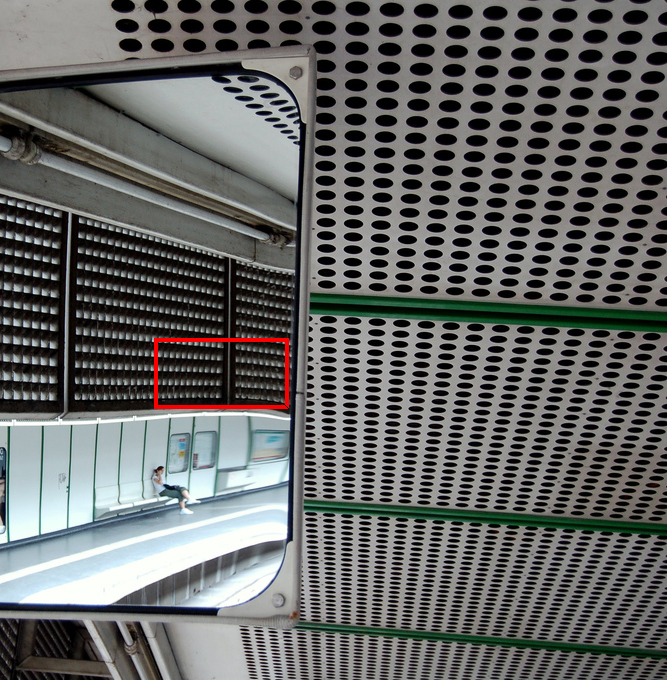}
	\\
	Ground-truth HR
	\\
	\textsc{Urban100}: img004
	\end{tabular}
	\end{adjustbox}
	\hspace{-0.46cm}
	\begin{adjustbox}{valign=t}
	\begin{tabular}{ccccc}
	\includegraphics[width=0.15\textwidth]{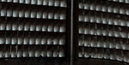} \hspace{\fs} &
	\includegraphics[width=0.15\textwidth]{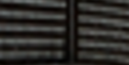} \hspace{\fs} &
	\includegraphics[width=0.15\textwidth]{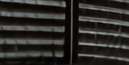} \hspace{\fs} &
	\includegraphics[width=0.15\textwidth]{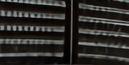} \hspace{\fs} &
	\includegraphics[width=0.15\textwidth]{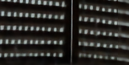} \hspace{\fs} 
    \vspace{1mm}
	\\
	HR \hspace{\fs} &
	Bicubic \hspace{\fs} &
	EDSR~\cite{lim2017edsr} \hspace{\fs} &
	RCAN~\cite{zhang2018rcan} \hspace{\fs} &
	CSNLN~\cite{Mei2020CSNLN} \hspace{\fs} 


	\vspace{1mm}
	\\
	\includegraphics[width=0.15\textwidth]{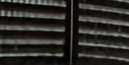} \hspace{\fs} &
	\includegraphics[width=0.15\textwidth]{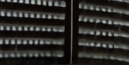} \hspace{\fs} &
	\includegraphics[width=0.15\textwidth]{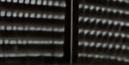} \hspace{\fs} &
	\includegraphics[width=0.15\textwidth]{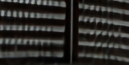} \hspace{\fs} &
	\includegraphics[width=0.15\textwidth]{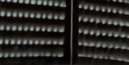}  \hspace{\fs} 
    \vspace{1mm}
	\\ 
	HAN~\cite{niu2020HAN} \hspace{\fs} &
	NLSN~\cite{Mei_2021_NLSN} \hspace{\fs} &
	SwinIR~\cite{liang21swinir} \hspace{\fs} &
	ENLCN~\cite{xia2022ENLCA} \hspace{\fs} &
	ITSRN++(ours) \hspace{\fs} 

	\\

	\\
	\end{tabular}
	\end{adjustbox}
	\vspace{2mm}
	\\
	\hspace{-0.4cm}
	\begin{adjustbox}{valign=t}
	\begin{tabular}{c}
	\includegraphics[width=0.2\textwidth]{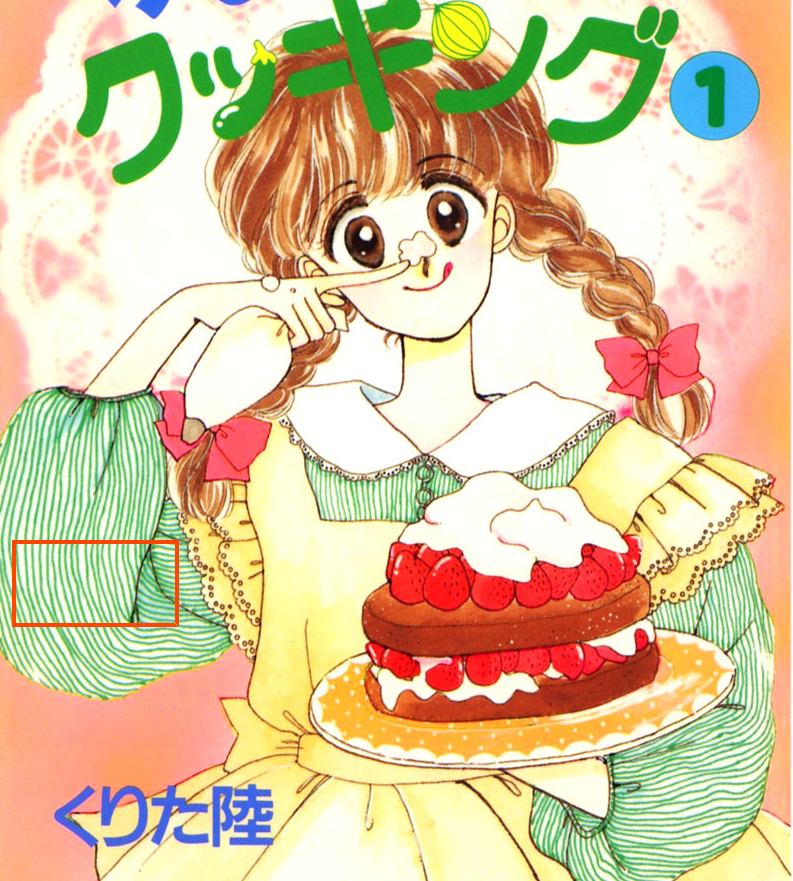}
	\\
	Ground-truth HR
	\\
	\textsc{Manga109}: YumeiroCooking
	\end{tabular}
	\end{adjustbox}
	\hspace{-0.46cm}
	\begin{adjustbox}{valign=t}
	\begin{tabular}{ccccc}
	\includegraphics[width=0.15\textwidth]{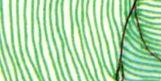} \hspace{\fs} &
	\includegraphics[width=0.15\textwidth]{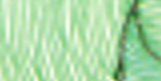} \hspace{\fs} &
	\includegraphics[width=0.15\textwidth]{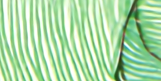} \hspace{\fs} &
	\includegraphics[width=0.15\textwidth]{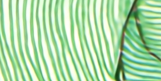} \hspace{\fs} &
	\includegraphics[width=0.15\textwidth]{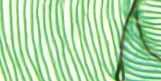} \hspace{\fs} 
    \vspace{2mm}
	\\
	HR \hspace{\fs} &
	Bicubic \hspace{\fs} &
	EDSR~\cite{lim2017edsr} \hspace{\fs} &
	RCAN~\cite{zhang2018rcan} \hspace{\fs} &
	CSNLN~\cite{Mei2020CSNLN} \hspace{\fs} 
 \vspace{2mm}
	\\

	\includegraphics[width=0.15\textwidth]{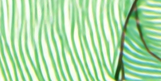} \hspace{\fs} &
	\includegraphics[width=0.15\textwidth]{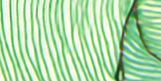} \hspace{\fs} &
	\includegraphics[width=0.15\textwidth]{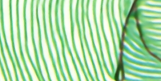} \hspace{\fs} &
	\includegraphics[width=0.15\textwidth]{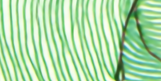} \hspace{\fs} &
	\includegraphics[width=0.15\textwidth]{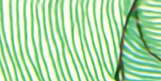}  \hspace{\fs} 
    \vspace{2mm}
	\\ 
	SwinIR~\cite{liang21swinir} \hspace{\fs} &
	SwinIR-MetaSR~\cite{liang21swinir,hu2019metasr} \hspace{\fs} &
	SwinIR-LIIF~\cite{liang21swinir,chen2021liif} \hspace{\fs} &
	SwinIR-LTE~\cite{liang21swinir, lee2021lte} \hspace{\fs} &
	ITSRN++(ours) \hspace{\fs} 

    \\

	\end{tabular}
	\end{adjustbox}
	
	\vspace{2mm}
	\end{tabular}
	\caption{
		\textbf{Visual comparison for $\times4$ SR on the \textsc{Urban100} and \textsc{Manga109} datasets.}
	}
	\label{fig:result_4x_natural}
\end{figure*}

\subsection{Comparisons with state-of-the-arts} 

We compare the proposed method with eight state-of-the-art SR algorithms, including single-scale SR methods, \textit{i.e.,} EDSR~\cite{lim2017edsr}, RDN~\cite{zhang2018RDN}, RCAN~\cite{zhang2018rcan}, SwinIR~\cite{liang21swinir}, and arbitrary-scale SR methods, \textit{i.e.}, MetaSR\cite{hu2019metasr}, LIIF\cite{chen2021liif}, LTE\cite{lee2021lte}, and our conference approach ITSRN~\cite{yang2021itsrn}. \textcolor{red}{Since the original MetaSR did not provide the implementations for large scale ($> 4\times$) upsampling, we re-implement its large upsampling according to ~\cite{chen2021liif}.} 
Following the default setting of their codes, LIIF, ITSRN, and LTE use RDN \cite{zhang2018RDN} as the feature extraction backbone. Since single-scale SR methods rely on specific up-sampling module, they need to train different models for different upsampling scales and cannot be tested for the scales not in the training process. Therefore, we only compare with them on $\times2, \times3, \times4$ SR. Besides evaluating on screen content image SR, we also evaluate the SR performance on natural images to demonstrate the effectiveness of the proposed ITSRN++.

\subsubsection{Comparisons on SCI SR}
All the compared methods are retrained on the training set of the proposed SCI2K dataset and evaluated on the testing set of SCI2K and three 
benchmark screen content datasets, \textit{i.e.}, SCID, CCI, and SIQAD. 
%
\tabref{quality_val} presents the quantitative comparisons on the test set of our SCI2K dataset. It can be observed that our method consistently outperforms all the compared methods in terms of both PSNR and SSIM. Specifically, our method ITSRN++ outperforms the second best method SwinIR by 0.74 dB for $\times 3$ SR. Meanwhile, our method is much better than the compared continuous SR methods. One reason is that our upsampler is better than theirs and the other reason is that our feature extraction backbone is better than RDN.  
%
\tabref{quality_benchmark} further presents the SR results on three SCI quality assessment datasets. We directly utilize the SR models trained on SCI2K to test. It can be observed that our method still outperforms the compared methods. Our gains over the second best method (SwinIR) on the three datasets are larger than that on the SCI2K test set. It demonstrates that our method has better generalization ability than SwinIR. The visual comparisons on the four screen content test sets are presented in \figref{result_4x_sci}. It can be observed that our method recovers more sharp edges and realistic characters than the compared methods. In summary, ITSRN++ achieves the best quantitative and qualitative results on screen content image SR. 

In addition, we evaluate the SR performance when the LR images are compressed. Table \ref{table:comp} presents the comparison results. All the models are retrained on the training set of SCI2K-compression. The test sets are processed by JPEG  compression with quality factors set to 75, 85, and 95 respectively. It can be observed that our method still outperforms all the compared methods.  

\begin{figure}[h]
	\centering
	\footnotesize
	\resizebox{0.48\textwidth}{!}{
			\begin{tabular}{c}
				\includegraphics[width=1\textwidth]{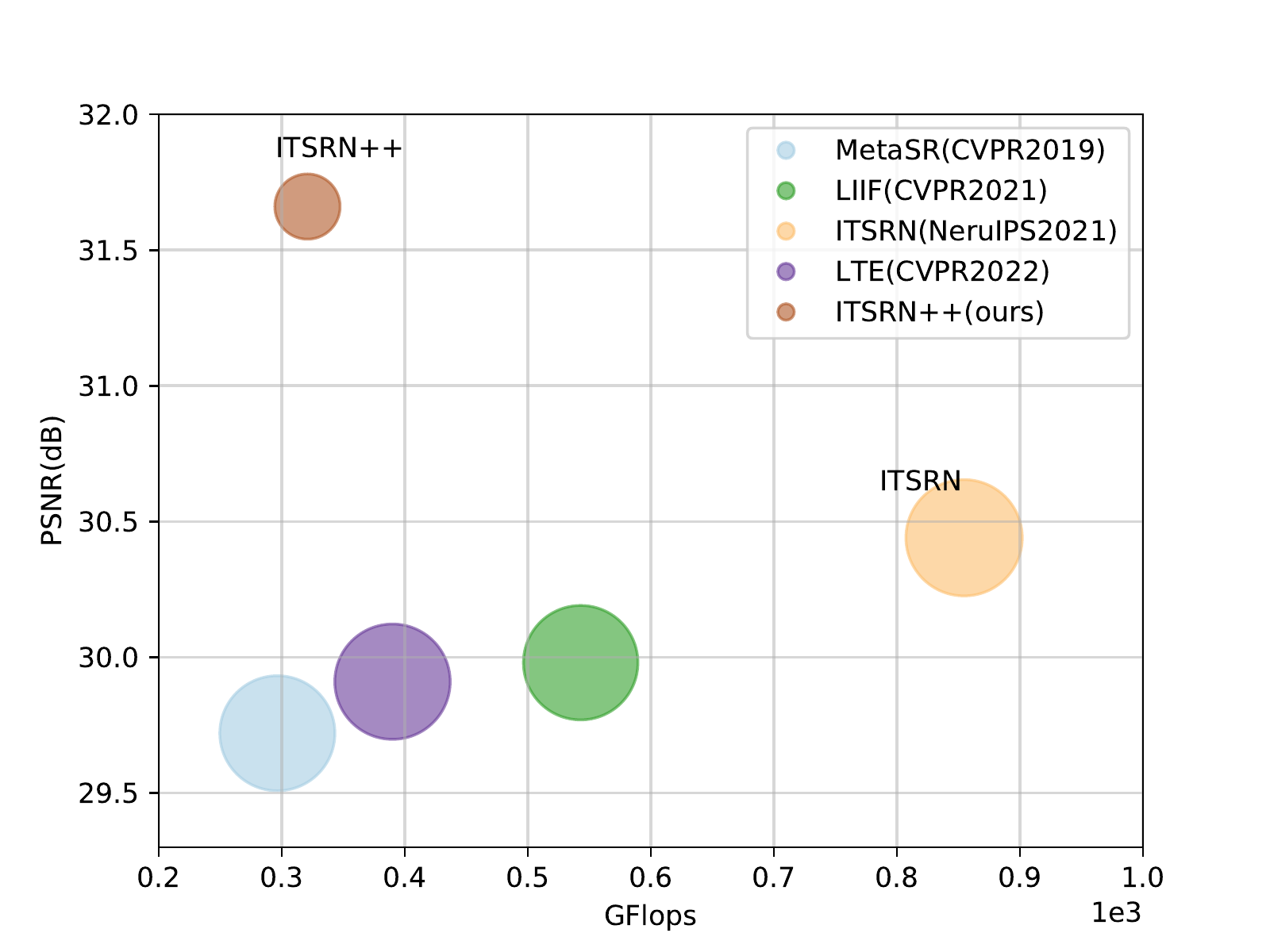}
				\\[1ex]
			\end{tabular}
		}
	\caption{Computing complexity comparison. 
The PSNR results are evaluated on the SCI2K test set for $\times4$ SR. The FLOPs are calculated with a $128 \times 128$ input. The circle size indicates the number of parameters.}
	\label{fig:psnr_parameter}
\end{figure}
\figref{psnr_parameter} presents the performance versus the FLOPs of continuous SR methods. The number of parameters is represented by the size of the circle. A larger circle indicate a larger number of parameters. \textcolor{red}{Compared with our previous version ITSRN, ITSRN++ saves about 60\% FLOPs, but brings nearly 1 dB gain.} Comparing to the MetaSR~\cite{hu2019metasr} and LIIF~\cite{chen2021liif}, our ITSRN++ has over 1 dB gain for $4\times$ SR.



\subsubsection{Comparisons on Natural Image SR}
Since most SR methods are evaluated on natural images, we further compare with state-of-the-arts on natural image SR. We re-train our method with DIV2K dataset \cite{agustsson2017div2k} and evaluate on five  benchmark datasets, \textit{i.e.}, \textsc{Set5}~\cite{bevilacqua2012set5}, \textsc{Set14}~\cite{zeyde2010set14}, \textsc{BSDS100}~\cite{martin2001b100}, \textsc{Urban100}~\cite{huang2015urban100}
and \textsc{Mange109}~\cite{matsui2017manga109}. The compared methods are also trained on DIV2K and their results are directly quoted from the corresponding papers. Note that, the results for SwinIR-MetaSR and SwinIR-LIIF are realized by LTE \cite{lee2021lte}.

The quantitative results are presented in~\tabref{quality_benchmark_natual}.
It can be observed that our ITSRN++ consistently outperforms existing methods on \textsc{Urban100} and \textsc{Mange109} datasets, since the images in the two datasets have many sharp edges (\textsc{Urban100} is constructed by urban buildings and \textsc{Mange109} is constructed by manga, one kind of screen contents). For example, for $\times4$ SR, ITSRN++ outperforms SwinIR-LTE (the second best method) by 0.23 dB and 0.22 dB on \textsc{Manga109} and \textsc{Urban100},  respectively. Meanwhile, our FLOPs are smaller than those of SwinIR-LTE. This demonstrates that the 
point-wise modulated upsampler and dual branch block are beneficial for sharp edge reconstruction. On the other three test sets, our method is comparable or slightly better than the compared methods. This verifies that our method can also work well for natural image SR. 
%
%
We also test on $\times 8$ SR to evaluate the continuity and robustness of our model. Note that, our ITSRN++ do not "see"  $\times 8$ pairs in training, but we still achieve the best performance in four benchmarks except \textsc{Set5}. For example, compared with SwinIR which is trained with $\times 8$ LR-HR pairs, our method still achieves 0.32 dB gain on \textsc{Manga109}. Compared with the second best method (SwinIR-LTE), our method achieves 0.16 dB gain.   

Fig. \ref{fig:result_4x_natural} presents the visual comparison results for $\times4$ SR on the \textsc{Urban100} and \textsc{Manga109} datasets. Our method accurately reconstructs parallel straight lines, grid patterns, and texts. 
\section{Conclusions and Limitation}
In this work, we propose a better and stronger implicit transformer network for screen content image SR. With the proposed modulation based implicit transformer for upsampler and the enhanced explicit transformer for feature extraction, the proposed method achieves more than 1 dB gain against our previous conference version ITSRN. Experiments on four screen content image datasets and five benchmark natural image SR datasets demonstrate the superiority and generalizability of our method. Besides, a large high resolution screen content image dataset~\textit{SCI2K} is constructed, which will benefit the development of SCI SR methods. 

We would like to point out that continuous SR methods usually consume more computing resources compared with fixed scale SR methods. Since there are MLPs in our upsampler, its computing complexity is higher than pixel shuffle based upsampler. In the future, we would like to optimize the upsampler to make it lighter.


 
%
%

\section*{Acknowledgment}
This research was supported in part by the National Natural Science Foundation of China under Grant 62072331 and Grant 62231018.





\bibliographystyle{IEEEtran}
\bibliography{reference}
\end{document}